\documentclass[11pt,leqno]{article}
\usepackage[T1]{fontenc}
\usepackage{amsmath,amssymb,amsfonts,amsthm}
\usepackage{mathrsfs}
\usepackage{stmaryrd}
\usepackage[margin=1in]{geometry}

\usepackage{graphicx}
\usepackage{bm}
\usepackage{booktabs}
\usepackage{array}

\usepackage{natbib}
\usepackage{xcolor}
\usepackage{hyperref}
\swapnumbers
\newtheorem{definition}{Definition}[section]
\newtheorem{proposition}{Proposition}[section]

\newtheorem{lemma}{Lemma}[section]

\newcommand{\Cl}{\mathrm{Cl}}
\newcommand{\grade}[1]{\langle #1 \rangle}
\newcommand{\sem}[1]{\llbracket #1 \rrbracket}
\newcommand{\lc}{\,\lrcorner\,}
\newcommand{\rc}{\,\llcorner\,}

\newcommand{\Rev}[1]{\widetilde{#1}}

\title{\textbf{Toward a Functional Geometric Algebra \\ for Natural Language Semantics}}
\author{James Pustejovsky\\
Computer Science Department\\
Brandeis University\\
\texttt{jamesp@brandeis.edu}}
\date{}

\begin{document}
\maketitle

\begin{abstract}
\noindent Distributional and neural approaches to natural language semantics have been built almost exclusively on conventional linear algebra: vectors, matrices, tensors, and the operations that accompany them. These methods have achieved remarkable empirical success, yet they face persistent structural limitations in compositional semantics, type sensitivity, and interpretability. I argue in this paper that \emph{geometric algebra} (GA) -- specifically, Clifford algebras -- provides a mathematically superior foundation for semantic representation, and that a \emph{Functional Geometric Algebra} (FGA) framework extends GA toward a typed, compositional semantics capable of supporting inference, transformation, and interpretability while retaining full compatibility with distributional learning and modern neural architectures. I develop the formal foundations, identify three core capabilities that GA provides and linear algebra does not, present a detailed worked example illustrating operator-level semantic contrasts, and show how GA-based operations already implicit in current transformer architectures can be made explicit and extended. The central claim is not merely increased dimensionality but increased \emph{structural organization}: GA expands an $n$-dimensional embedding space into a $2^n$ multivector algebra where base semantic concepts and their higher-order interactions are represented within a single, principled algebraic framework.
\end{abstract}

% ============================================================
\section{Introduction}

Distributional and neural approaches to natural language semantics have largely been built on conventional linear algebra (LA): vectors, matrices, tensors, and operations such as addition, inner products, and linear transformations. 
These methods have achieved impressive empirical success, from static embeddings \citep{mikolov2013,pennington2014} through contextual representations \citep{devlin2019,radford2019} to today's large language models. As \citet{erk2012} documents in her survey of vector space models for word and phrase meaning, the field has been remarkably successful at capturing lexical similarity but faces persistent and well-documented challenges when the question turns from word meaning to compositional phrase meaning -- challenges that motivate the entire framework developed here:

\begin{itemize}
\item \emph{Weakly structured composition.} Vector addition, concatenation, and bilinear maps do not intrinsically encode the \emph{mode} of semantic combination. Whether a verb combines with its subject, an adjective modifies a noun, or a quantifier scopes over a predicate, the algebraic operation is the same: a learned matrix multiplication. The compositional structure must be entirely induced from data. This is a curious situation for a field that has, since Frege, understood meaning to be compositional.

\item \emph{Parameter inflation for higher-order interactions.} Modeling interactions beyond pairwise similarity -- ternary relations, event structure, scopal ambiguity -- requires either deeper networks, additional tensor parameters, or explicit architectural mechanisms (e.g., multi-head attention). The interaction structure is implicit in weights rather than explicit in the algebra.

\item \emph{Limited intrinsic support for types.} Natural language meaning is fundamentally typed: entities, predicates, propositions, events, and modifiers differ in kind, not just in direction within a shared vector space. Standard LA provides no native mechanism for maintaining type distinctions; these must be imposed externally through architectural choices. Anyone working in formal semantics will immediately recognize the cost: without types, there is no principled way to say why \emph{John sleeps} is well-formed while \emph{*Sleeps John very} is not.

\item \emph{Difficulty integrating symbolic and geometric structure.} Formal semantic theories provide compositional rules, variable binding, quantification, and inference, but operate over discrete structures. Distributional models provide continuous similarity and gradient-based learning, but lack logical structure. Bridging the two has proven notoriously difficult within the confines of standard LA.
\end{itemize}

I argue in this paper that \emph{geometric algebra} (GA), and specifically the Clifford algebras $\Cl(V, Q)$, provides a more expressive mathematical substrate for semantic representation that addresses all four limitations simultaneously. Moreover, a \emph{Functional Geometric Algebra} (FGA) framework extends GA toward a typed, compositional semantics in which lexical meaning, compositional structure, and contextual modulation are all expressed within a single algebra.

Geometric algebra and related algebraic extensions of vector spaces have been explored for NLP in a number of ways. \citet{widdows2003} demonstrated that the subspace lattice of a word-vector space, equipped with orthogonal complements and subspace sums, the connectives of quantum logic,  outperforms Boolean operators for word-sense discrimination and document retrieval, providing early evidence that the algebraic structure of semantic vector spaces extends well beyond the inner product. 
\citet{pilato2007} used GA rotors for sub-symbolic encoding of sentence structure within an LSA semantic space, and \citet{mani2016} proposed multivector word embeddings in which the grade structure captures a semantic hierarchy. In knowledge graph embeddings, \citet{xu2020} and \citet{demir2023} have demonstrated that the geometric product can effectively model relational patterns. What I propose here extends these precedents from encoding and embedding into a full compositional semantics, providing formal operations for predicate application, role binding, and type coercion within a unified algebraic framework.

Independently, the machine learning community has begun exploring neural architectures built directly on Clifford-algebraic structure. \citet[GATr;][]{brehmer2023} develop GATr as a transformer architecture in projective geometric algebra, showing that multivector-based computation can be implemented at transformer scale. \citet{hirst2026}  extend this direction with Versor, a conformal-geometric sequence architecture built around rotor-based state updates, reporting strong performance together with substantial parameter efficiency.
These results provide direct empirical evidence, albeit in neighboring domains, that the Clifford algebra is not merely a theoretical elegance but a practical architectural advantage. FGA proposes to bring this same algebraic substrate to natural language semantics, where the linguistic motivations for graded structure, compositional binding, and type coercion are even more explicit.

The argument of this paper proceeds as follows. I begin in Section~2 by introducing the mathematical foundations of geometric algebra and its relationship to linear algebra, with an emphasis on making the semantic motivations transparent at each step. Section~3 addresses the expressivity-versus-efficiency concern -- a natural worry when one first encounters the $2^n$-dimensional multivector algebra. Section~4 develops the graded structure of Clifford algebras as a natural semantic type hierarchy. Section~5 works out the compositional mechanics in detail: typed domains, lexical entries, application via contraction, structure building via the wedge product, and type checking via grade arithmetic. Section~6 presents three further capabilities that GA provides for semantics -- including broader coverage of concepts, polysemy, and modification. Section~7 develops the FGA framework and its three-layer architecture. Section~8 works through three compositional sketches of increasing richness -- verb--argument composition, qualia-sensitive modification, and complement coercion -- showing specifically where LA fails and FGA succeeds. Section~9 connects GA to existing neural architectures, showing that key GA operations are already implicit in current systems. Section~10 compares FGA systematically with alternative compositional frameworks. Section~11 develops a plausible path toward neural implementation, including multivector parameterization, type-sensitive learning, and regularization, and Section~12 concludes.

% ============================================================
\section{From Linear Algebra to Geometric Algebra}

\subsection{What linear algebra gives NLP -- and what it leaves out}

Standard embedding models represent lexical items as vectors $\mathbf{v} \in \mathbb{R}^n$. The basic operation for measuring semantic similarity is the inner (dot) product:
\begin{equation}
a \cdot b = \sum_i a_i b_i \;\in\; \mathbb{R}.
\end{equation}
The result is a scalar: a single number encoding how well two meanings ``align.'' This is extremely useful -- it powers cosine similarity, attention scores, and essentially all neural retrieval -- but it is also \emph{lossy}. Two vectors that have the same dot product with a third can nevertheless point in very different directions. The dot product collapses the full geometric relationship between two vectors into a single number, discarding all information about \emph{how} and \emph{in what plane} they differ.

Composition is even more limited. Vector addition ($a + b$), concatenation ($[a; b]$), and learned bilinear maps ($a^\top W b$) are the standard toolkit, but none of these operations intrinsically encodes what \emph{kind} of composition is being performed. Whether a verb combines with its subject, an adjective modifies a noun, or a quantifier scopes over a predicate, the algebraic operation is the same: a learned matrix multiplication. The compositional mode must be entirely induced from data. This is, I think, a significant limitation -- and it is one that geometric algebra directly addresses.

I should note explicitly that this limitation is not specific to static embeddings like Word2Vec or GloVe, whose inadequacy for compositionality is well established and was a principal motivation for the move to contextual models. The limitation I am identifying is structural: it concerns the \emph{vector space substrate itself}. BERT and GPT represent meanings as vectors in $\mathbb{R}^n$ and compose them through learned matrix multiplications and attention operations -- exactly the same algebraic toolkit, applied contextually rather than statically. Contextualization addresses the problem of fixed word representations but does not address the deeper problem that the algebra over those representations has no native notion of binding, type, oriented composition, or recoverable structure. The compositional limitations of flat vector spaces persist even when the vectors are contextualized, and it is these algebraic limitations that FGA targets.

\subsection{The geometric product: similarity and structure in one operation}

For two vectors $a, b$ in a vector space $V$, the geometric product decomposes as:
\begin{equation}
ab = \underbrace{a \cdot b}_{\text{scalar (grade 0)}} + \underbrace{a \wedge b}_{\text{bivector (grade 2)}}.
\end{equation}
The first term, $a \cdot b$, is the familiar symmetric inner product -- the same scalar similarity that LA provides. The second term, $a \wedge b$, is the \emph{wedge} (or \emph{outer}) product: an antisymmetric quantity ($a \wedge b = -b \wedge a$) that represents the \emph{oriented plane} spanned by $a$ and $b$. Its magnitude is the area of the parallelogram formed by the two vectors; its sign encodes which vector comes first. If $a$ and $b$ are parallel, $a \wedge b = 0$; if they are orthogonal, $a \wedge b$ is maximal. The wedge product captures exactly the directional and relational information that the dot product discards.

The key insight is that the geometric product $ab$ keeps \emph{both pieces of information simultaneously}, in a single algebraic object. This is not a concatenation or a tuple -- it is a structured element of an algebra in which these components interact under multiplication.

\paragraph{A note on addition in GA.} A reader familiar with Hyper-Dimensional Computing or Vector Symbolic Architectures may wonder what ``addition'' means in this context, since a multivector like $\alpha + \mathbf{v} + B$ appears to ``add'' objects of different kinds (a scalar, a vector, a bivector). The answer is that this is not bundling in the VSA sense. In HDC/VSA, adding two hypervectors produces a noisy superposition in the \emph{same} space, and similarity checks degrade as more items are bundled. In GA, addition across grades is a \emph{direct sum}: the scalar, vector, and bivector components live in algebraically orthogonal subspaces of the Clifford algebra, and no information is lost or blurred by combining them into a single multivector. Each grade component can be cleanly extracted by grade projection $\langle M \rangle_k$ at any time. The ``addition'' is formal bookkeeping -- a way of packaging distinct pieces of information into one algebraic object -- not a lossy superposition. Where VSA bundling is inherently noisy, GA addition is exact.

\subsection{Clifford algebras: the formal framework}

I will adopt standard definitions and notations from the major classical  treatments of geometric algebra throughout the paper \citep{hestenes1984,doran2003,dorst2007}. 

\begin{definition}[Clifford Algebra]
Let $V$ be a real vector space of dimension $n$ equipped with a quadratic form $Q: V \to \mathbb{R}$. The \emph{Clifford algebra} $\Cl(V, Q)$ is the associative algebra generated by $V$ subject to the fundamental relation:
\begin{equation}
v^2 = Q(v) \quad \text{for all } v \in V.
\end{equation}
For an orthonormal basis $\{e_1, \ldots, e_n\}$, this entails:
\begin{equation}
e_i e_j = -e_j e_i \quad (i \neq j), \qquad e_i^2 = Q(e_i).
\end{equation}
\end{definition}

The quadratic form $Q$ determines a \emph{signature} $(p, q, r)$ where $p$ basis vectors square to $+1$, $q$ to $-1$, and $r$ to $0$, with $p + q + r = n$. It is worth pausing to explain what this means, since the signature is central to FGA's expressive power. The ``negative'' generators -- the $q$ basis vectors satisfying $e_i^2 = -1$ -- are not ``negative dimensions'' in any spatial sense. They are directions along which the quadratic form is \emph{negative-definite}: squaring a vector along such a direction yields $-1$ rather than $+1$. The analogy with Minkowski spacetime is apt and worth making explicit: in $\Cl(1,3,0)$, the familiar metric $t^2 - x^2 - y^2 - z^2$ is precisely the quadratic form, where the timelike dimension squares to $+1$ while the three spatial dimensions square to $-1$, and the sign difference is what encodes the causal structure of spacetime. In FGA, we can exploit mixed signatures similarly: positive-squaring generators might encode spatial, extensional, or similarity-preserving features, while negative-squaring generators encode intensional, modal, or contrastive features. The null generators ($r$ basis vectors squaring to $0$) are useful for projective constructions, as in the projective geometric algebra $\Cl(3,0,1)$ used by GATr \citep{brehmer2023}. The signature is thus a design parameter of the semantic space, not merely a mathematical technicality.

This design parameter has  algebraic consequences that I will exploit throughout the paper. If $\mathbf{v}$ lives in a positive-squaring subspace ($\mathbf{v}^2 = +|\mathbf{v}|^2$) and $\mathbf{u}$ lives in a negative-squaring subspace ($\mathbf{u}^2 = -|\mathbf{u}|^2$), then the bivector $\mathbf{v} \wedge \mathbf{u}$ formed from one vector in each subspace squares to a \emph{positive} value: $(\mathbf{v} \wedge \mathbf{u})^2 = -\mathbf{v}^2\mathbf{u}^2 = -(+|\mathbf{v}|^2)(-|\mathbf{u}|^2) = +|\mathbf{v}|^2|\mathbf{u}|^2$. By contrast, a bivector formed from two positive-squaring vectors squares \emph{negatively}: $(\mathbf{v}_1 \wedge \mathbf{v}_2)^2 = -|\mathbf{v}_1|^2|\mathbf{v}_2|^2 < 0$. This sign difference determines the character of the rotor generated by the bivector: negative-squaring bivectors generate \emph{elliptic rotors} (smooth, cyclic rotations), while positive-squaring bivectors generate \emph{hyperbolic rotors} (non-cyclic boosts that stretch or compress along an axis). In Section~\ref{sec:broader}, I will show that this distinction maps directly onto the semantic distinction between extensional modification (cyclic, similarity-based) and intensional modification (non-cyclic, function-oriented).

The two equations in the definition above have an intuitive explanation. Equation~(3), $v^2 = Q(v)$, says that squaring a vector in the Clifford algebra yields a scalar -- intuitively, the ``self-similarity'' of a direction. This is what ties the algebra to the geometry of the underlying space. Equation~(4), $e_i e_j = -e_j e_i$ for $i \neq j$, says that the product of two different basis vectors is \emph{anticommutative}. This anticommutativity is the algebraic source of orientation: it is what makes the wedge product antisymmetric, what distinguishes $a \wedge b$ from $b \wedge a$, and ultimately what allows GA to encode argument order intrinsically. Together, these two properties -- scalar self-product and anticommutative cross-product -- generate the entire Clifford algebra from the vector space and its quadratic form.

A familiar example: $\Cl(3, 0, 0)$ generates the algebra over 3D Euclidean space. The quaternions arise as the even subalgebra $\Cl^+(3, 0, 0)$: the three unit bivectors $e_2 e_3$, $e_3 e_1$, $e_1 e_2$ each square to $-1$, anticommute pairwise, and -- crucially -- their products close on one another, e.g.\ $(e_2 e_3)(e_3 e_1) = -e_1 e_2$. This closure is what makes them isomorphic to Hamilton's $\{i, j, k\}$; with an appropriate choice of signs (e.g.\ $i = e_3 e_2$, $j = e_1 e_3$, $k = e_2 e_1$) one recovers the relation $ij = k$.
(A reader more accustomed to thinking of quaternions via $\Cl(0, 2, 0)$, where the generators $e_i$ themselves square to $-1$, will find the same algebra by a different route; the isomorphism $\Cl^+(3,0,0) \cong \Cl(0,2,0)$ makes these viewpoints formally equivalent, and both are useful intuition pumps for the idea that algebraic structure can encode geometric relationships.) For a first semantic application, the Euclidean signature $\Cl(n, 0, 0)$ preserves familiar similarity structure while gaining the full graded algebra; pseudo-Euclidean signatures $\Cl(p, q, 0)$ may encode semantic asymmetries such as defeasibility or contrast.

The quaternion example above is deliberately simple, and a reader encountering this material for the first time might form the impression that FGA is an application of quaternions, which it is not. Quaternions are a four-dimensional subalgebra ($\Cl^+(3,0,0) \cong \Cl(0,2,0)$) that handles rotations in 3D and nothing else. The full Clifford algebra $\Cl(V, Q)$ is far richer: it includes all grades from 0 to $n$, supports arbitrary dimension and mixed signature, and provides not only rotors but also contraction (function application), the wedge product (binding), grade projection (type extraction), and exact unbinding -- none of which quaternions offer. The quaternion example is included because it connects the unfamiliar algebra to a familiar object; but the FGA framework uses the full Clifford algebra, of which quaternions are a  special case.

\subsection{Grades: a built-in type system}

The most important structural property of a Clifford algebra for semantic purposes is its \emph{graded} decomposition. Any multivector $M \in \Cl(V,Q)$ decomposes uniquely as a sum of grade projections:
\begin{equation}
M = \sum_{k=0}^{n} \langle M \rangle_k, \qquad \langle M \rangle_k \in \Cl^k(V, Q), \quad \dim \Cl^k(V, Q) = \binom{n}{k}.
\end{equation}

We write $\langle M \rangle_k$ for the \emph{grade-$k$ projection} of $M$ -- the operation that extracts just the grade-$k$ component. This is analogous to extracting the scalar or vector part of a complex expression; it will be a central notational tool throughout this paper.

To make the graded structure clear, consider the smallest nontrivial example. For $n = 4$ basis vectors $\{e_1, e_2, e_3, e_4\}$, the full Clifford algebra $\Cl(4, 0, 0)$ has $2^4 = 16$ basis elements organized by grade:

\begin{center}
\begin{tabular}{clcll}
\toprule
\textbf{Grade} & \textbf{Basis elements} & \textbf{Count} & \textbf{Geometric object} & \textbf{Semantic analogue} \\
\midrule
0 & $1$ (the unit scalar) & $\binom{4}{0} = 1$ & Scalar & Truth value \\
1 & $e_1,\; e_2,\; e_3,\; e_4$ & $\binom{4}{1} = 4$ & Vectors & Entity, property \\
2 & $e_1 e_2,\; e_1 e_3, \ldots$ & $\binom{4}{2} = 6$ & Bivectors (planes) & Binary relation \\
3 & $e_1 e_2 e_3, \ldots$ & $\binom{4}{3} = 4$ & Trivectors (volumes) & Event frame \\
4 & $e_1 e_2 e_3 e_4$ & $\binom{4}{4} = 1$ & Pseudoscalar & Discourse frame \\
\bottomrule
\end{tabular}
\end{center}

A general element of $\Cl(4, 0, 0)$ -- a \emph{multivector} -- is a linear combination of all 16 basis elements:
\begin{equation}
M = \underbrace{\alpha}_{\langle M \rangle_0} + \underbrace{v_1 e_1 + \cdots + v_4 e_4}_{\langle M \rangle_1} + \underbrace{B_{12}\, e_1 e_2 + \cdots + B_{34}\, e_3 e_4}_{\langle M \rangle_2} + \underbrace{T_{123}\, e_1 e_2 e_3 + \cdots}_{\langle M \rangle_3} + \underbrace{\omega\, e_1 e_2 e_3 e_4}_{\langle M \rangle_4}.
\end{equation}
The grade structure is not a notational artifact -- it is algebraically enforced. The geometric product of two grade-1 vectors always produces a grade-0 scalar plus a grade-2 bivector. The wedge product of three vectors always produces a grade-3 trivector. And so on. This is what I mean when I say GA has a ``built-in type system'': the grade of an algebraic expression is determined by the grades of its inputs and the operation applied, just as the type of a semantic expression is determined by the types of its constituents and the mode of composition.

\paragraph{A terminological note.} The term ``multivector'' as used throughout this paper refers exclusively to the Clifford-algebraic sense defined above: a single element of $\Cl(V,Q)$ composed of components at multiple grades. This should not be confused with the unrelated use of ``multi-vector'' in the information retrieval literature, where it refers to representing a document as a \emph{set} of separate vectors (one per token), as in ColBERT-style late-interaction models \citep{khattab2020}. In that setting, ``multi-vector'' is a collection of standard vectors; in our setting, a multivector is a single algebraic object with internal grade structure. The two concepts share a surface term but have entirely different mathematical content.

\paragraph{Blades.} A related term that appears throughout the paper is \emph{blade}. A $k$-blade is a multivector that can be expressed as the outer (wedge) product of $k$ vectors: $B = v_1 \wedge v_2 \wedge \cdots \wedge v_k$. It represents a $k$-dimensional oriented subspace of the vector space. Every blade is a \emph{homogeneous} multivector -- it has a single definite grade $k$ -- but not every grade-$k$ multivector is a blade (a sum of two bivectors in different planes, for instance, may not be factorizable as a single wedge product). Blades are the ``pure'' elements of each grade, and they play an important role in FGA: role--filler bindings are bivector blades ($r \wedge f$), verb templates are $k$-blades in role space, and concepts can be modeled as blades representing oriented semantic subspaces (Section~\ref{sec:broader}).

The rightmost column of the table above hints at why this matters for linguistics, and I want to draw the reader's attention to it before we proceed to the operations. In formal semantics, meaning is organized by \emph{type}: entities ($e$), truth values ($t$), predicates ($e \to t$), relations ($e \to e \to t$), and so on. The grade hierarchy provides a direct algebraic encoding of this type structure. A grade-0 scalar is the natural home for a truth value -- the end result of fully saturating a predicate with its arguments. A grade-1 vector represents an entity or a one-place predicate. A grade-2 bivector represents a binary relation: its two ``legs'' correspond to the two argument positions, and its orientation encodes which is which (agent vs.\ patient, for instance). The key consequence is operational: \emph{contracting} a grade-2 relation with a grade-1 entity yields a grade-1 result -- a unary predicate with one argument saturated -- mirroring the type rule $(e \to e \to t) \times e \to (e \to t)$. The algebra does the type bookkeeping automatically. This correspondence is developed formally in Section~4 and worked out compositionally in Section~5.

\subsection{Four operations and their semantic analogues}

Four operations on multivectors are fundamental to the FGA framework. I introduce each alongside its semantic motivation, since the linguistic payoff is immediate in every case.

\paragraph{(1) The inner product} ($a \cdot b = \langle ab \rangle_0$) extracts the scalar (grade-0) part of the geometric product. This is the familiar dot product -- semantic similarity. In FGA, it also serves as the mechanism for \emph{full} predicate application: when a unary predicate contracts with its argument, the result is a truth value (a scalar).

\paragraph{(2) The wedge product} ($a \wedge b = \langle ab \rangle_2$ for vectors) extracts the bivector (grade-2) part. It is antisymmetric: $a \wedge b = -b \wedge a$, so it intrinsically encodes argument order. Semantically, it serves as \emph{role--filler binding}: the bivector $r_{\textsc{agent}} \wedge \mathbf{john}$ binds the agent role to its filler, and the antisymmetry ensures that the role and filler are distinguished. This is the GA analogue of Smolensky's tensor product variable binding \citep{smolensky1990}, but within a bounded algebra rather than an ever-expanding tensor space.

\paragraph{(3) Left contraction} ($a \lc B$) reduces the grade of $B$ by the grade of $a$, with $a$ acting from the left. Formally, for a grade-$j$ element $A$ and a grade-$k$ element $B$ with $j \leq k$:
\begin{equation}
A \lc B = \langle AB \rangle_{k-j}.
\end{equation}
This is the algebraic analogue of \emph{function application}: a grade-1 vector (an argument) contracting into a grade-2 bivector (a binary relation) yields a grade-1 vector (a unary predicate with one argument saturated).

\paragraph{(4) Right contraction} ($B \rc a$) is the mirror of left contraction: it reduces the grade of $B$ by the grade of $a$, but with $a$ acting from the right:
\begin{equation}
B \rc A = \langle BA \rangle_{k-j}.
\end{equation}
Left and right contraction are not equivalent -- $a \lc B \neq B \rc a$ in general -- and this asymmetry is semantically crucial. For a binary relation $\ell \in \Cl^2$, right-contracting with the object $m$ saturates one argument position ($\ell \rc m \in \Cl^1$), while left-contracting the result with the subject $j$ saturates the other ($j \lc (\ell \rc m) \in \Cl^0$). The two contractions encode the two distinct argument slots of the predicate, and the algebra keeps track of which is which. This directly encodes argument directionality without requiring external categorical structure.

\paragraph{(5) The rotor sandwich} ($x' = R\, x\, \Rev{R}$) transforms a multivector $x$ by ``sandwiching'' it between a rotor $R$ and its reverse $\Rev{R}$. A rotor is an even-grade element satisfying $R\Rev{R} = 1$. The sandwich product is:
\begin{itemize}
\item \textbf{Grade-preserving}: if $x$ is a vector, $x'$ is a vector; if $x$ is a bivector, $x'$ is a bivector. Ontological categories are maintained.
\item \textbf{Norm-preserving}: $|x'| = |x|$. The ``magnitude'' of the concept is not distorted.
\item \textbf{Invertible}: $x = \Rev{R}\, x'\, R$. The original meaning is recoverable.
\item \textbf{Composable}: applying rotor $R_1$ then $R_2$ is equivalent to applying $R_2 R_1$. Chains of transformations compose algebraically.
\end{itemize}
Semantically, rotors model \emph{type coercion and contextual modulation}: rotating a representation within its grade subspace to shift its interpretation without changing its ontological category. The classic Generative Lexicon example -- \emph{begin the book} coercing ``book'' toward its telic quale ``reading'' -- is modeled as a rotor acting on the book's vector.

\subsection{GA vs.\ LA: a side-by-side comparison}

It may be helpful at this point to step back and see the overall picture. The following table summarizes how standard NLP operations map between the two frameworks:

\begin{center}
\renewcommand{\arraystretch}{1.25}
\begin{tabular}{@{}p{3.5cm}p{4.5cm}p{5cm}@{}}
\toprule
\textbf{Semantic operation} & \textbf{Linear algebra} & \textbf{Geometric algebra} \\
\midrule
Similarity & Dot product $a \cdot b \to$ scalar & Inner product $\langle ab \rangle_0 \to$ scalar \\
Role--filler binding & Tensor product $r \otimes f$ (unbounded rank) & Wedge product $r \wedge f$ (bounded grade) \\
Argument recovery & Approximate (SVD, cleanup) & Exact via contraction: $r^{-1} \lc (r \wedge f) = f$ \\
Function application & Learned matrix $Wx$ & Left contraction $a \lc B$ (grade reduction) \\
Type coercion & Implicit in contextual embeddings & Rotor sandwich $Rx\Rev{R}$ (explicit, invertible) \\
Type checking & No native mechanism & Grade comparison (intrinsic) \\
\bottomrule
\end{tabular}
\end{center}

The central point of this comparison is not that GA \emph{adds operations} to LA, but that it \emph{unifies} them. The inner product, wedge product, and contraction are all aspects of a single operation -- the geometric product -- decomposed by grade projection. It is this algebraic unity that makes a single-framework approach to compositional semantics possible, and it is the reason I believe the framework deserves serious attention.

\section{Expressivity Without Parameter Inflation}

A natural objection presents itself immediately: moving from $n$ dimensions to $2^n$ basis elements looks computationally prohibitive. I want to address this concern directly, because the answer is both important and, I think, somewhat surprising. The key counterargument is structural:

\begin{quote}
\emph{Geometric algebra does not add arbitrary parameters; it organizes interactions systematically within a closed algebraic structure.}
\end{quote}

Let us examine the situation in specific terms. A conventional embedding model with $n = 10$ places each word in $\mathbb{R}^{10}$. Modeling pairwise interactions between dimensions requires learning separate parameters -- a bilinear model needs an $n \times n$ weight matrix (100 parameters), and modeling three-way interactions requires an $n \times n \times n$ tensor (1000 parameters). Each level of interaction multiplies the parameter count.

The Clifford algebra $\Cl(10, 0, 0)$ has $2^{10} = 1024$ basis elements organized by grade: 1 scalar, 10 vectors, 45 bivectors, 120 trivectors, and so on. Critically, these 1024 dimensions are not free parameters in the sense of a flat vector: the bivector $e_1 e_2$ is the antisymmetric product of $e_1$ and $e_2$, and their \emph{interaction rules} are algebraically determined by the geometric product. In a learned multivector embedding, the coefficients of each grade are free parameters during training, but the operations that compose two multivectors -- the way bivector components of one representation interact with vector components of another -- are constrained by the algebra. The savings come at composition time, not at representation time.

This means that a GA embedding carries structured higher-order information ``for free'' -- the bivector component of a word's multivector representation encodes relational information, the trivector component encodes three-way interactions, and so on, all within a single algebraic object rather than requiring separate parameter tensors. The result is a representation that is informationally richer than a flat vector of the same total dimensionality, precisely because its internal structure constrains and organizes how the components interact under algebraic operations.

A practical note: for neural implementations, one need not represent the full $2^n$-dimensional multivector. Grade-truncated representations that retain only grades 0 through $k$ (for small $k$) capture the most semantically relevant structure while keeping dimensionality manageable. For a base dimension of $n = 64$ (comparable to small embedding models), retaining grades 0--2 gives $1 + 64 + 2016 = 2081$ components -- substantial, but far less than the full $2^{64}$, and rich enough to encode entities, predicates, and binary relations within a single algebraic object.

To state the parameter advantage directly: a grade-$k$-truncated multivector in $n$ dimensions has $\sum_{j=0}^{k} \binom{n}{j}$ components. The corresponding LA representation of an object with $k$-ary relational structure requires a rank-$k$ tensor with $n^k$ parameters. For $n = 300$ (a standard embedding dimension) and $k = 2$ (binary relations), the FGA representation has $1 + 300 + 44{,}850 = 45{,}151$ parameters per item, compared to $300^2 = 90{,}000$ for a tensor. At $k = 3$ (ternary event structures), the FGA count is approximately $4.5$ million versus $27$ million for the tensor. The advantage grows with arity: FGA's parameter count grows as $O(n^k/k!)$ (via the binomial coefficient), while the tensor count grows as $O(n^k)$ without the $k!$ compression. This compression is not a trick; it reflects the algebraic fact that the wedge product encodes relational structure antisymmetrically, eliminating the redundant degrees of freedom that a symmetric tensor must represent explicitly.

% ============================================================
\section{Graded Structure as Semantic Type Hierarchy}

The grade-to-type correspondence previewed in Section~2.4 is not merely an analogy -- it is algebraically enforced. Because the grade decomposition $M = \sum_k \langle M \rangle_k$ is maintained under every algebraic operation -- every product of multivectors produces a result whose grade is determined by the grades of its inputs -- the type structure is intrinsic to the algebra rather than externally imposed. The idea of exploiting grade structure for semantic purposes has precedent: \citet{mani2016} proposed that grades could encode a part-of-speech hierarchy in multivector word embeddings, with smaller grades for more specific concepts and larger grades for more general ones, and trained a Word2Mvec model using the geometric product in a skip-gram architecture. What I propose here extends this intuition from an empirical embedding strategy to a formal compositional framework, in which the grade-to-type mapping is motivated by the type-theoretic structure of natural language semantics and the algebraic operations respect the type distinctions. The following table makes the full correspondence explicit:

\begin{center}
\begin{tabular}{lll}
\toprule
\textbf{Grade} & \textbf{Geometric Object} & \textbf{Semantic Interpretation} \\
\midrule
0 & Scalar & Truth values, salience, certainty \\
1 & Vector & Entities, basic predicates \\
2 & Bivector & Binary relations, thematic role bindings \\
3 & Trivector & Ternary relations, event frames \\
$k$ & $k$-blade & $k$-ary structures \\
$n$ & Pseudoscalar & Discourse orientation, modal perspective (see \S\ref{sec:broader}) \\
\bottomrule
\end{tabular}
\end{center}

What matters here is that these type distinctions are \emph{intrinsic to the algebra}. In standard LA, a vector representing an entity and a vector representing a predicate live in the same space and are distinguished only by convention or by network architecture. In GA, they occupy different grades, and the algebraic operations respect these distinctions: the contraction of a grade-1 entity vector into a grade-2 relational bivector yields a grade-1 result, mirroring the type-theoretic rule that applying a binary relation to one argument yields a unary predicate.

I want to emphasize that this is not merely a notational convenience. It provides what I would call \emph{algebraic inductive bias}: the grade structure constrains what operations are well-formed and what results they produce, reducing the space of possible compositions that a model must search during learning. When composition respects grade structure, the algebra enforces a form of type discipline that LA-based models must learn entirely from data.

A candid remark is in order about the status of this correspondence. I am proposing the grade-type mapping as an \emph{inductive bias} -- a structural hypothesis that organizes the representational space and constrains the learning problem -- not as a claim that learned representations will decompose crisply into grade-0 truth values, grade-1 entities, grade-2 relations, and so on. The history of distributed representations suggests that the actual geometry of learned embeddings is typically messier than any a priori scheme predicts: BERTology studies \citep{rogers2021} have shown that transformer representations encode syntactic and semantic information in ways that are real and recoverable but not neatly localized into the dimensions or layers that one might expect from first principles. I would expect the same to hold for learned multivector representations: the grade structure provides a scaffold, but the learned representations will spread across grades in ways that reflect the statistical structure of the data as much as the theoretical type hierarchy. What the grade-based inductive bias buys is not a guarantee that the correspondence will be crisp, but a constraint that makes it more likely to emerge than it would in a flat vector space -- and, critically, a set of algebraic operations (contraction, wedge product, rotor action) that \emph{respect} grade structure even when the representations are not perfectly grade-aligned. The right empirical question is not ``Do entities live exclusively at grade 1?'' but ``Do grade-truncated representations with grade-sensitive composition outperform flat representations of equivalent dimensionality on compositional generalization tasks?'' That question is testable, and I believe it is the one that matters.

% ============================================================
\section{Compositional Mechanics: From Types to Derivations}

The grade-to-type correspondence introduced above is not merely a labeling scheme -- it drives a fully compositional semantics in which lexical entries, predicate application, argument structure, and type checking all have specific algebraic realizations. Let me now develop this machinery in some detail, since it forms the operational core of FGA.

\subsection{Typed domains}

I make the type-to-grade mapping precise with the following assignments:
\begin{align}
e &\mapsto \Cl^1(V) && \text{(individuals as vectors)}, \\
t &\mapsto \Cl^0(V) && \text{(truth values as scalars)}, \\
e \to t &\mapsto \Cl^1(V) && \text{(unary predicates as vectors)}, \\
e \to (e \to t) &\mapsto \Cl^2(V) && \text{(binary relations as bivectors)}, \\
n\text{-place predicate} &\mapsto \Cl^n(V) && \text{(}n\text{-ary predicates as }n\text{-vectors)}.
\end{align}
Two observations are immediate. First, both entities ($e$) and unary predicates ($e \to t$) are grade-1 vectors, but they are distinguished by their \emph{subspace support}. To make this precise, I introduce the notion of a direct sum decomposition of the base vector space. We write $V_1 \oplus V_2$ for the \emph{direct sum} of two subspaces: this is the space of all elements $v_1 + v_2$ with $v_i \in V_i$, subject to the constraint that $V_1 \cap V_2 = \{0\}$ -- i.e., the subspaces share no nonzero elements. The direct sum ensures \emph{non-interference}: components in $V_1$ and components in $V_2$ are independently accessible and do not contaminate each other under projection. This is the algebraic mechanism that keeps typed domains separate within a single algebra.

Let the base semantic space decompose as $V_E \oplus V_P \oplus \cdots$, where $V_E$ is the entity subspace and $V_P$ is the predicate-content subspace. Entities have support only in $V_E$: $j \in \Cl^1(V_E)$. Unary predicates have support in both $V_E$ and $V_P$: $R \in \Cl^1(V_E \oplus V_P)$. The $V_E$ component of a predicate encodes selectional information -- what kinds of entities it applies to -- and is the component that participates in contraction with entity arguments. The $V_P$ component encodes the predicate's own semantic content. Application (contraction) exploits the shared $V_E$ component: $R \cdot j = R_E \cdot j + R_P \cdot j = R_E \cdot j \in \Cl^0$, since $V_P \perp V_E$ ensures $R_P \cdot j = 0$. The distinction between entities and predicates is thus algebraically intrinsic -- inspectable by subspace projection -- not a matter of external annotation. This mirrors the Montagovian situation where both $e$ and $t$ are basic types, distinguished by their domains. Second, the arity of a predicate is directly encoded in its grade: an $n$-place predicate requires $n$ contractions to reach grade~0 (a truth value), each contraction consuming one argument.

\subsection{Lexical entries}

Lexical items receive multivector representations whose grades reflect their semantic types. The examples that follow are deliberately simple -- the point is to show that the type-to-grade mapping produces well-behaved compositional behavior even in the most basic cases.

\paragraph{Entities.}
Proper nouns and definite descriptions denote grade-1 vectors in the entity subspace:
\begin{align}
\sem{\text{John}} &= j \in \Cl^1(V_E), \quad \sem{\text{Mary}} = m \in \Cl^1(V_E), \quad \sem{\text{the book}} = b \in \Cl^1(V_E).
\end{align}
Throughout, entity vectors are assumed to be \emph{unit-normalized}: $|j| = |m| = |b| = 1$. This ensures that the inner product $j \cdot m = \cos\theta$ is a cosine similarity with a constant identity value ($j \cdot j = 1$ for all entities), and that rotor action preserves norm. Normalization is not merely a convention but an algebraic commitment: it ensures that the scalar part of any geometric product involving entity vectors has a stable, interpretable scale.\footnote{A reader may object that $\sem{\text{the book}}$ is treated here as a pre-composed entity rather than being derived compositionally from the determiner \emph{the} and the common noun \emph{book}. This is a simplification for expository purposes. In a full treatment, determiners are operators -- grade-preserving or grade-modifying transformations -- that compose with common-noun predicates to produce entity-type referential expressions. The mechanism is analogous to type-raising: the definite article shifts a predicate-type common noun into an entity-type expression by contracting with a uniqueness presupposition. Demonstratives (\emph{this}, \emph{that}) would involve rotors that additionally encode spatial or discourse-deictic orientation. These compositional details are orthogonal to the present argument and are developed in \citet{pustejovsky2026fga}.}

\paragraph{Unary predicates.}
Intransitive verbs, common nouns, and one-place adjectives denote grade-1 vectors with support in both the entity and predicate-content subspaces:
\begin{align}
\sem{\text{run}} &= R \in \Cl^1(V_E \oplus V_P), \quad \sem{\text{human}} = H \in \Cl^1(V_E \oplus V_P), \quad \sem{\text{happy}} = S \in \Cl^1(V_E \oplus V_P).
\end{align}
The inner product of a predicate's $V_E$ component with an entity vector yields a scalar (grade-0), corresponding to truth-conditional evaluation. Common nouns additionally carry qualia rotors; adjectives carry modification-specific profiles (cf.\ Section~6.4).

\paragraph{Binary relations.}
Transitive verbs denote grade-2 bivectors with support in the entity and predicate-content subspaces:
\begin{align}
\sem{\text{love}} &= \ell \in \Cl^2(V_E \oplus V_P), \quad \sem{\text{kick}} = K \in \Cl^2(V_E \oplus V_P), \quad \sem{\text{read}} = \rho \in \Cl^2(V_E \oplus V_P).
\end{align}
As bivectors, transitive verbs are \emph{oriented planes} -- the orientation encodes which argument position is which. The anticommutativity of the algebra ensures that $\sem{\text{love}}(j, m) \neq \sem{\text{love}}(m, j)$, because contracting from different sides of a bivector yields different results.

\subsection{Application via contraction}

Predicate application -- the most fundamental compositional operation in any semantics -- is realized in FGA by the inner product (contraction), which reduces grade by one per argument. This mirrors the type-theoretic rule that applying a function to an argument eliminates one layer of the function type. The correspondence is exact, and worth seeing in detail.

\paragraph{Unary application.}
For a unary predicate $P \in \Cl^1$ and an entity $x \in \Cl^1$:
\begin{equation}
\sem{P(x)} = P \cdot x \in \Cl^0 \quad \text{(scalar/truth value)}.
\end{equation}
The contraction $\Cl^1 \times \Cl^1 \to \Cl^0$ corresponds to the type reduction $(e \to t) \times e \to t$.

\paragraph{Example: \emph{Mary laughs}.}
$\sem{\text{laugh}(\text{Mary})} = L \cdot m \in \Cl^0$. The result is a scalar whose magnitude encodes the degree to which the predication holds -- in a Boolean model, this is 0 or 1; in a graded model, it is a continuous value reflecting plausibility or distributional support.

\paragraph{Curried application for transitive verbs.}
A binary relation $\ell \in \Cl^2$ takes two arguments via successive contractions -- but now the directionality matters. Right contraction applies the object from the right; left contraction applies the subject from the left. The first application is partial:
\begin{equation}
\ell \rc m = \ell_m \in \Cl^1 \quad \text{(``love Mary'' as a unary property)},
\end{equation}
reducing the bivector to a vector by saturating the object position. The second application saturates the subject:
\begin{equation}
j \lc (\ell \rc m) \in \Cl^0 \quad \text{(``John loves Mary'' as a truth value)}.
\end{equation}
The contraction sequence $\Cl^2 \xrightarrow{\rc\, m} \Cl^1 \xrightarrow{j\, \lc} \Cl^0$ mirrors the curried type reduction $e \to (e \to t)$ applied successively to two arguments. Grade decreases with each argument saturation, and the distinction between left and right contraction encodes which argument position is being filled -- a distinction that the symmetric inner product of standard LA cannot express.

\subsection{Outer Application and Inner Application}
\label{sec:oaia}

The contraction mechanism developed above -- which I will call \emph{Outer Application} (OA) -- is the classical case of function-argument composition: the predicate takes its argument as an undifferentiated whole, and contraction reduces grade accordingly. In the Montagovian tradition, this corresponds to function application; in classical vector-space semantics, it corresponds to matrix-vector multiplication.

But composition in natural language is almost never this simple. The vast majority of predicates do not take their arguments as undifferentiated wholes. ``Wipe the table'' targets the table's surface, not the table as a whole. ``Tie your shoe'' accesses the laces, a specific subpart meeting physical conditions for being tied. ``Begin the book'' accesses the book's characteristic function and begins that. ``Good knife'' evaluates the knife along its cutting function, not along some absolute quality scale. In each case, the predicate \emph{selects into} the argument's internal structure, composing not with the whole but with a specific aspect of it.

In the theory of Distributed Compositionality, developed recently within Generative Lexicon theory \citep{pustejovskyJezek2026}, this selective mode of composition is called \emph{Inner Application} (IA). In FGA, Inner Application is realized as a two-step process: first, a quale-targeting rotor transforms the argument's multivector, foregrounding the relevant subspace component; then contraction saturates the argument position with the rotor-transformed representation:
\begin{equation}
\sem{P(\text{arg})} = P \lc \big(R_Q \; \sem{\text{arg}} \; \Rev{R}_Q\big),
\end{equation}
where $R_Q$ is a rotor encoding quale $Q$ (Telic, Constitutive, Agentive, or Formal). The rotor is what Inner Application \emph{is} in the Clifford algebra: it is the formal mechanism by which a predicate probes the internal organization of its argument and extracts the aspect it needs. When no quale-targeting rotor is active -- when $R_Q$ is the identity -- Inner Application reduces to Outer Application, and the derivation is classical function application. This ensures that FGA is conservative over standard compositional semantics.

The central claim is that Inner Application is the \emph{general} case, and Outer Application is the \emph{special} case. Predicates like ``exist'' or ``disappear,'' which genuinely want the full, undifferentiated argument, are the exception. The default mode of verb-argument composition involves a rotor -- a quale-mediated probe into the argument's subspace structure -- before contraction applies. Coercion in ``begin the book'' is not a special mechanism triggered by type mismatch; it is Inner Application with a cross-subspace rotor (from $V_E$ to $V_P$). Constitutive projection in ``wipe the table'' is Inner Application with an intra-subspace rotor (within $V_E$). Subsective modification in ``good knife'' is Inner Application accessing the Telic quale. The mechanism is uniform; what varies is the geometric target of the rotor:

\begin{itemize}
\item \textbf{Outer Application} (contraction alone, no rotor): the predicate takes the argument as an undifferentiated whole. This is the mode of predicates like ``exist,'' ``disappear,'' or ``arrive.''
\item \textbf{Inner Application, intra-subspace} (rotor within $V_E$): the predicate accesses a constitutive component of the argument -- its surface (``wipe''), its contents (``drink''), a specific subpart (``tie'').
\item \textbf{Inner Application, cross-subspace} (rotor from $V_E$ to $V_P$ or $V_S$): the predicate accesses a functional or eventive component of the argument -- its Telic quale (``begin the book,'' ``good knife''), its Agentive quale (``bake the cake'').
\end{itemize}

This inversion of the traditional framing -- IA as default, OA as special case -- has consequences for the architecture developed in Section~\ref{sec:architecture}: most actual composition is a Layer~3 $\to$ Layer~1 sequence (rotor then contraction), not Layer~1 alone. I illustrate this through the compositional sketches in Section~\ref{sec:sketches}.

\subsection{Structure building via the wedge product}

While the inner product \emph{consumes} arguments (grade reduction), the wedge product \emph{builds} structure (grade increase). This duality is one of the most attractive features of the framework: the same algebra that evaluates predications also constructs the event structures that predications are about.

Before developing the mechanics, a brief orientation may be helpful for readers less familiar with event-based semantics. In truth-conditional semantics (the tradition from Frege and Montague), the meaning of a sentence like \emph{John kicked the ball} is ultimately a truth value -- a scalar in FGA terms. But linguistic analysis also requires a richer representation that makes the \emph{participants} and \emph{structure} of the described situation explicit. In Davidsonian and neo-Davidsonian event semantics \citep{davidson1967,parsons1990}, a sentence denotes a structured \emph{event} $e$ with participants bearing \emph{thematic roles}: $\text{Agent}(e) = \text{John}$, $\text{Theme}(e) = \text{ball}$. The event representation captures ``who did what to whom'' in a way that a bare truth value cannot. In FGA, we represent thematic roles as designated vectors in a role subspace $V_R$: $r_{\textsc{ag}}$ for the agent role, $r_{\textsc{th}}$ for the theme role, and so on. We write $\mathcal{E}$ for an event representation -- a structured multivector encoding the event's participants and their roles. The wedge product provides the binding mechanism, and contraction provides the querying mechanism, as I now develop.

In the decomposed space $V = V_R \oplus V_E$, role--entity binding produces bivectors:
\begin{equation}
r \wedge e \in \Cl^1(V_R) \wedge \Cl^1(V_E) \subset \Cl^2(V).
\end{equation}
Each binding is an oriented plane spanned by a role direction and an entity direction, and the anticommutativity $r \wedge e = -e \wedge r$ ensures that the role and filler are distinguished.

An event is represented as a \emph{sum} of such bindings, not a single higher-grade blade:
\begin{equation}
\mathcal{E}_{\text{kick}} = (r_{\textsc{ag}} \wedge j) + (r_{\textsc{th}} \wedge b) \in \Cl^2(V).
\end{equation}
This avoids the associativity complications that arise from wedging the bindings together (which would produce a grade-4 element conflating the event's structure).

\subsection{Querying and unbinding}

The binding story above has an algebraic dual: a structured system of \emph{querying} operations that recover fillers, roles, and nested structure from bound representations. This dual is one of FGA's sharpest advantages over alternative binding architectures, and it deserves careful development.

\paragraph{The vector inverse.}
Every nonzero vector $v$ in a Clifford algebra has an exact inverse:
\begin{equation}
v^{-1} = \frac{v}{|v|^2},
\end{equation}
and more generally, for a nonzero blade $B$ of any grade (recall from Section~2.4 that a $k$-blade is the wedge product of $k$ vectors, representing an oriented $k$-dimensional subspace):
\begin{equation}
B^{-1} = \frac{\Rev{B}}{|B|^2},
\end{equation}
where $\Rev{B}$ is the \emph{reverse} of $B$ (its factors written in reverse order). This inversion is exact -- it introduces no statistical noise and does not degrade as a function of the number of items superposed in the ambient multivector. For normalized vectors ($|v| = 1$), the inverse is simply the vector itself: $v^{-1} = v$.

\paragraph{Basic unbinding.}
Given a binding $B = r \wedge f$, the filler is recovered exactly by left contraction with the inverse of the role:
\begin{equation}
r^{-1} \lc B = r^{-1} \lc (r \wedge f) = f.
\end{equation}
Symmetrically, the role is recovered by right contraction with the inverse of the filler:
\begin{equation}
B \rc f^{-1} = (r \wedge f) \rc f^{-1} = r.
\end{equation}
Both recoveries are algebraically exact. This is in direct contrast to HRR, where retrieval error grows as $O(1/\!\sqrt{n})$ for $n$ superposed bindings \citep{plate1995}.

\paragraph{Asymmetric querying.}
Because the wedge product is anticommutative ($r \wedge f = -(f \wedge r)$), the two contraction directions are not equivalent:
\begin{equation}
r^{-1} \lc B \neq B \rc r^{-1}.
\end{equation}
The left contraction $r^{-1} \lc B$ answers: \emph{given the role, what is the filler?} The right contraction $B \rc f^{-1}$ answers: \emph{given the filler, what is the role?} These are genuinely distinct semantic operations, and the algebra distinguishes them automatically. For linguistic applications the distinction is immediate: ``What did John see?'' (filler query given an agent) and ``What role does Mary play in this event?'' (role query given a filler) are geometrically distinct operations, not merely the same operation with different inputs. In HRR, both queries employ the same circular correlation\footnote{In Plate's HRR \citep{plate1995}, \emph{binding} is circular convolution ($\circledast$) and \emph{unbinding} is circular correlation (convolution with the approximate inverse $\widetilde{r}$). The point here is that HRR's unbinding operation -- correlation -- is the same operation regardless of whether one is querying for a filler given a role or a role given a filler; there is no algebraic distinction between the two query directions.} -- there is no geometric differentiation between the two query directions.

\paragraph{Querying superposed structures.}
An event record is a superposition of bindings:
\begin{equation}
\mathcal{E} = \sum_i (r_i \wedge f_i).
\end{equation}
To query this superposition for the filler of role $r_i$:
\begin{equation}
\mathrm{query}(\mathcal{E}, r_i) = r_i^{-1} \lc \langle \mathcal{E} \rangle_2.
\end{equation}
If the role vectors form an orthonormal set -- which is achievable by construction -- then $r_i^{-1} \lc (r_j \wedge f_j) = \delta_{ij}\, f_j$, so each contraction returns exactly $f_i$ and zero for all other bindings. Cross-binding crosstalk is eliminated \emph{algebraically}, not statistically. Compare the analogous HRR retrieval:
\begin{equation}
\widetilde{r_i} \circledast \sum_j (r_j \circledast f_j) = f_i + \underbrace{\sum_{j \neq i} (\widetilde{r_i} \circledast r_j) \circledast f_j}_{\text{noise term}},
\end{equation}

\noindent 
where the noise term is only approximately zero in expectation for random role vectors \citep{plate2003}. FGA makes what HRR approximates exact.

\paragraph{Grade projection as depth filter.}
Before applying contraction, one can filter by \emph{binding depth} using grade projection $\langle M \rangle_k$. Grade-1 components are unbound lexical items; grade-2 components are binary bindings (role--filler pairs); grade-3 components are ternary structures (full argument frames); and so on. This yields a navigable hierarchy of compositional depth that HRR cannot replicate, since all HRR bindings collapse to the same dimensionality. Grade projection is computationally inexpensive: in the standard basis representation, grade-$k$ components occupy a contiguous block of $\binom{n}{k}$ coefficients, so projection reduces to block selection with no arithmetic overhead.

\paragraph{Grade descent for nested structures.}
One of the most linguistically important query patterns is nested event retrieval. Consider a grade-3 blade encoding ``Agent claims Content,'' where the content is itself a bound event. Querying for the content returns not a filler vector but a blade of lower grade, which can itself be queried recursively. Each successive contraction reduces grade by one, navigating down the compositional hierarchy. The grade of the result at each step signals how much compositional structure remains to be unpacked -- a direct algebraic analogue of the recursive unbinding that formal semantics requires for embedded clauses.

\subsection{A full derivation: \emph{John loves Mary}}

To show that these pieces fit together, let me work through a complete derivation, producing both the truth-conditional and event-structural representations of a transitive sentence from the same lexical inventory.

\paragraph{Lexical inventory.}
$j, m \in \Cl^1(V_E)$ (entities); $\ell \in \Cl^2(V_E \oplus V_P)$ (predicate content of \emph{love}); $r_{\textsc{ag}}, r_{\textsc{th}} \in \Cl^1(V_R)$ (role vectors).

\paragraph{Truth-conditional meaning} (via successive contraction in $V_E$):
\begin{equation}
\sem{\text{John loves Mary}}_t = j \lc (\ell \rc m) \in \Cl^0.
\end{equation}

\paragraph{Event structure} (via wedge-product binding in $V_R \oplus V_E$):
\begin{equation}
\sem{\text{John loves Mary}}_e = (r_{\textsc{ag}} \wedge j) + (r_{\textsc{th}} \wedge m) \in \Cl^2(V).
\end{equation}

\paragraph{Recovery test:}
$r_{\textsc{ag}}^{-1} \lc \mathcal{E} = j$, \quad $r_{\textsc{th}}^{-1} \lc \mathcal{E} = m$. Both fillers are recovered exactly, as developed in Section~5.5.

The two representations -- truth-conditional and event-structural -- are computed within the same algebra using different operations: the inner product for evaluation, the wedge product for structure. This is the two-layer design that Section~\ref{sec:architecture} develops into the full three-layer FGA architecture. The important point is that no additional mechanisms were needed: everything followed from the algebra and the typed lexical entries.

\subsection{Verbs as role templates}

A verb's argument structure can be represented independently from its predicate content as a \emph{blade in role space} specifying which thematic roles it requires:
\begin{align}
\sem{\text{run}}_{\textsc{roles}} &\sim r_{\textsc{ag}} \in \Cl^1(V_R) && \text{(intransitive: needs agent)}, \\
\sem{\text{kick}}_{\textsc{roles}} &\sim r_{\textsc{ag}} \wedge r_{\textsc{th}} \in \Cl^2(V_R) && \text{(transitive: needs agent and theme)}, \\
\sem{\text{give}}_{\textsc{roles}} &\sim r_{\textsc{ag}} \wedge r_{\textsc{th}} \wedge r_{\textsc{goal}} \in \Cl^3(V_R) && \text{(ditransitive)}.
\end{align}
The grade of the role-space blade directly encodes the verb's valence. Application takes a role template and a list of arguments and expands it into a sum of role--entity bindings: $\text{Apply}(r_{\textsc{ag}} \wedge r_{\textsc{th}}, \langle j, b \rangle) = (r_{\textsc{ag}} \wedge j) + (r_{\textsc{th}} \wedge b)$. The role template thus functions as a combinatory specification -- a bridge between subcategorization and event structure.

\subsection{Type checking via grade arithmetic and subspace compatibility}

One of the more satisfying consequences of the grade-based type system is that it enforces well-typedness algebraically. But the enforcement mechanism has two components, not one, and it is important to state both precisely. The grade rules are:

\begin{center}
\begin{tabular}{ll}
Contraction: & $\text{grade}(a \lc B) = \text{grade}(B) - \text{grade}(a)$ \quad (when $a$ has support in $B$'s subspace) \\
Wedge product: & $\text{grade}(a \wedge b) = \text{grade}(a) + \text{grade}(b)$
\end{tabular}
\end{center}

Grade arithmetic provides a \emph{necessary} condition for well-typedness: a fully saturated predication must reach grade~0 (scalar). If the final grade is not~0, unsaturated arguments remain, and the grade tells you how many. But grade arithmetic alone is not \emph{sufficient} for type checking, as the following contrast reveals.

Consider legitimate partial application of a transitive verb: $\sem{\text{love}} \rc \sem{\text{John}} = \ell \rc j$, where $\ell \in \Cl^2$ and $j \in \Cl^1$. The result has grade $2 - 1 = 1$ -- a vector representing the unary property ``loves John,'' ready to be saturated by a subject. Now consider the ill-typed composition $\sem{\text{run}} \lc \sem{\text{love}}$, where $R \in \Cl^1$ and $\ell \in \Cl^2$. Grade arithmetic again gives $2 - 1 = 1$. Both results have grade~1. But the first is a well-formed residual predicate, and the second is a type error -- a unary predicate applied to a binary relation is semantically nonsensical.

What distinguishes them? The answer lies in the \emph{subspace support} of the contraction. The contraction $a \lc B$ is nonzero only when $a$ has support in the subspace spanned by $B$ -- informally, when $a$ can be ``factored out'' of $B$. This is not an external stipulation; it follows from the definition of the contraction in the Clifford algebra.

For $\ell \rc j$: John is an entity vector in $V_E$, and $\ell$ is a bivector whose two argument positions are spanned by entity-type directions. John has support in the relevant subspace, the contraction is nonzero, and the result is a well-formed grade-1 residual predicate.

For $R \lc \ell$: the vector $R = \sem{\text{run}}$ is a unary predicate with support in $V_E \oplus V_P$. But $R$ is not an \emph{entity filler} for one of $\ell$'s argument slots -- it is another predicate, living in the wrong subspace to serve as an argument. The contraction $R \lc \ell$ \emph{vanishes}: it yields the zero vector. The type error is signaled not by an anomalous grade but by the \emph{nullity} of the result. Attempting to factor a predicate out of a relation fails, algebraically, because the predicate does not inhabit the argument subspace of the relation.

The wedge product provides a confirming diagnostic. The wedge $\ell \wedge j$ yields a grade-$3$ element -- this is the event-structural binding discussed in Section~5.6, where wedge products build neo-Davidsonian event representations. But $R \wedge \ell$ also has grade $1 + 2 = 3$, and this trivector has no legitimate semantic interpretation in the FGA framework: no well-formed composition of a unary predicate with a binary relation should produce a ternary structure. The anomalous \emph{semantic type} of the wedge product -- a trivector with no role-filler interpretation -- is the structural signal of incompatibility.

The full type-checking mechanism thus has three layers:

\begin{enumerate}
\item \textbf{Grade arithmetic} (necessary condition): the final grade must be 0 for a saturated proposition, 1 for a unary residual, etc. If the grade is wrong for the intended composition, the derivation is ill-typed.
\item \textbf{Nonvanishing of contraction} (primary diagnostic): contraction yields zero when the argument lacks support in the predicate's subspace. This catches type errors that grade arithmetic misses, including the $\sem{\text{run}} \lc \sem{\text{love}}$ case above.
\item \textbf{Semantic coherence of the wedge product} (confirming diagnostic): if the wedge product of two expressions produces a multivector grade with no semantic interpretation (e.g., a trivector from a unary predicate and a binary relation), the composition is structurally ill-formed.
\end{enumerate}

\noindent This three-layer system turns type checking from an external metalinguistic judgment into an algebraic property of the derivation itself. Grade arithmetic provides the quick check; subspace compatibility provides the deep check; and the wedge product provides the structural confirmation. All three are intrinsic to the Clifford algebra -- no external type checker is required.

% ============================================================
\section{Three Core Capabilities for Semantics}
\label{sec:capabilities}

Having established the formal mechanics, I turn now to the broader question: what does geometric algebra give us for semantics that linear algebra does not? The strongest motivation lies in three capabilities that are rarely achieved simultaneously in conventional embedding models.

\subsection{Structured Composition via the Wedge Product}

The outer (wedge) product $a \wedge b$ produces an oriented bivector -- a directed plane element -- that encodes the \emph{relational structure} between $a$ and $b$, not merely their similarity. This is precisely what is needed for predicate--argument binding, role structure, and event composition.

Role--filler binding has a substantial history in connectionist and vector-symbolic approaches. \citet{smolensky1990} introduced Tensor Product Variable Binding, using the outer (tensor) product $\mathbf{r} \otimes \mathbf{f}$ to bind a role vector to a filler vector. While powerful, tensor products inflate dimensionality: each binding increases the tensor order, and unbinding is approximate. \citet{pollack1990} took a different approach to the same problem: Recursive Auto-Associative Memory (RAAM) uses learned compression via backpropagation to encode variable-sized tree structures into fixed-width vectors, achieving bounded dimensionality without an explicit binding operation. RAAM demonstrated that compositional structure can be learned in fixed-width representations, but the composition is opaque, encoded in network weights rather than algebraic operation,  and reconstruction is approximate. The approach anticipated the tension that motivates FGA: bounded representations are essential, but so is algebraic transparency.

Holographic Reduced Representations \citep{plate1995} and Vector Symbolic Architectures \citep{kanerva2009} addressed the dimensionality issue through circular convolution and related operations, but at the cost of approximate, noise-accumulating unbinding. The VSA program's foundational insight -- that distributed representations require a genuine algebra over vectors, not merely a vector space -- is one that FGA fully inherits; the argument developed here concerns which algebra best realizes that program, not whether such an algebra is needed. Recent work by \citet{larsson2025} pushes HRR-based architectures further toward typed compositional semantics, approximating the lambda calculus within the Semantic Pointer Architecture (SPA) in order to bridge Cooper's Type Theory with Records \citep{cooper2023} and Eliasmith's computational neuroscience framework.  Their results are instructive: they achieve a working question-answering model, but the circular convolution at the heart of their system remains commutative (so argument order is not intrinsically encoded), unbinding requires cleanup memories (because recovery is approximate), and the lambda calculus must be \emph{simulated} rather than natively supported. These are precisely the limitations that motivate the move to geometric algebra. The GA wedge product offers a middle path: like the tensor product, it produces an exact, structured binding (the antisymmetric bivector $r \wedge f$); unlike the tensor product, it remains within the bounded grade structure of the Clifford algebra (the result is a grade-2 element, not a separate tensor space); and unlike circular convolution, recovery via contraction is algebraically exact rather than approximate.

Consider encoding the thematic structure of a simple transitive clause. In LA, binding an agent to an event requires either concatenation (losing algebraic structure), tensor products (inflating dimensionality), or learned transformations (losing interpretability). In GA, we represent role--filler bindings directly:
\begin{equation}
\text{event} = (r_{\textsc{agent}} \wedge \mathbf{john}) + (r_{\textsc{patient}} \wedge \mathbf{ball}) + (r_{\textsc{action}} \wedge \mathbf{kick}),
\end{equation}
where $r_{\textsc{agent}}, r_{\textsc{patient}}, r_{\textsc{action}}$ are role key vectors and $\mathbf{john}, \mathbf{ball}, \mathbf{kick}$ are entity/predicate vectors. The antisymmetry of the wedge product ensures that $r \wedge f \neq f \wedge r$, encoding the asymmetry of role assignment. And fillers are exactly recoverable via contraction:
\begin{equation}
r_{\textsc{agent}}^{-1} \lc \text{event} = \mathbf{john}.
\end{equation}

This is not possible with inner products alone, which are symmetric and scalar-valued. The wedge product adds the oriented, relational dimension that natural language semantics requires. The earliest application of GA to natural language encoding is due to \citet{pilato2007}, who used Clifford algebra rotors to encode word order in sentence representations built from an LSA semantic space. FGA extends this precedent from positional encoding to full compositional role--filler binding with exact algebraic recovery.

\subsection{Type Coercion and Contextual Modulation via Rotors}

Rotors provide principled representations for semantic transformations such as type coercion, perspective shift, and contextual reinterpretation. A rotor $R \in \Cl(V, Q)$ acts on a multivector $x$ by the sandwich product:
\begin{equation}
x' = R\, x\, \Rev{R},
\end{equation}
which is (i) grade-preserving -- an entity remains an entity, a relation remains a relation; (ii) norm-preserving -- the ``magnitude'' of the concept is not distorted; and (iii) invertible -- the original meaning is recoverable. These properties make rotors well suited for a significant class of semantic transformations -- those that reinterpret a concept without changing its basic ontological category or semantic weight. Not all semantic operations have these properties: hedging and intensification may require non-norm-preserving maps, and nominalization (converting a predicate into an entity) requires grade change. For such cases, other GA operations -- projections, grade-part extraction, non-unitary transformations -- are available within the same algebra. Rotors model the structure-preserving core.

Consider the classic type coercion example from Generative Lexicon theory \citep{pustejovsky1995}: \emph{begin the book}. The verb \emph{begin} selects for an event, but \emph{book} denotes a physical object. The coercion to an event reading (begin reading/writing the book) can be modeled as a rotor that rotates the book's representation within the semantic space toward the event-compatible subspace, specifically exploiting the telic quale (the purpose of the book, i.e., reading):
\begin{equation}
R_{\textsc{telic}}\, \sem{\text{book}}\, \Rev{R}_{\textsc{telic}} \approx \sem{\text{read-book}}.
\end{equation}

In LA-based models, such coercions are handled implicitly through contextual embeddings -- BERT's representation of ``book'' in the context of ``begin'' differs from its representation in isolation. But the transformation is opaque: it lives inside billions of parameters with no interpretable algebraic structure. Rotors make the transformation explicit, decomposable, and compositional. A chain of coercions corresponds to a composition of rotors, and the net transformation has a clear geometric interpretation as a rotation through semantic space.

\paragraph{Where do rotors come from?} A reader accustomed to the random-initialization paradigm of HDC/VSA -- where the starting point is random hypervectors and structure is built up from them -- may reasonably wonder about the provenance of rotors. They are not random objects; they arise from three distinct sources, each algebraically grounded:

\begin{enumerate}
\item \emph{Algebraically}, as the geometric product of two unit vectors: $R = \mathbf{u}\mathbf{v}$. This rotor encodes the rotation from direction $\mathbf{u}$ to direction $\mathbf{v}$ in the plane they span, by an angle equal to twice the angle between them. When applied via the sandwich product $R\,x\,\Rev{R}$, it rotates any vector $x$ in that plane by the same angle.

\item \emph{Exponentially}, via the rotor exponential: $R = e^{-B\theta/2}$, where $B$ is a unit bivector (specifying the plane of rotation) and $\theta$ is the rotation angle. This is the Clifford-algebraic generalization of Euler's formula $e^{i\theta} = \cos\theta + i\sin\theta$, where the bivector $B$ plays the role of the imaginary unit $i$. The qualia rotors of Generative Lexicon theory -- $R_{\textsc{telic}}$, $R_{\textsc{agentive}}$, and so on -- are constructed this way: the bivector $B$ specifies \emph{which semantic plane} the coercion rotates in (e.g., the plane connecting the physical-object and reading-event regions of semantic space), and $\theta$ specifies \emph{how far} to rotate.

\item \emph{By learning}. In a neural implementation, the bivector $B$ and angle $\theta$ (or equivalently, the bivector coefficients $b_{ij}$) can be treated as trainable parameters optimized by gradient descent. This is how contextual rotors would be learned in an FGA-transformer: the network learns \emph{which rotation} to apply based on the compositional context, just as current models learn contextual embeddings -- but the learned object is a geometrically interpretable transformation rather than an opaque parameter update.
\end{enumerate}

The contrast with HDC/VSA is instructive. In VSA, binding vectors are typically random and carry no intrinsic geometric meaning; the system works \emph{despite} the arbitrariness of the initial encoding, because the algebraic properties of bundling and binding hold in expectation over random vectors. In FGA, rotors carry \emph{structured geometric meaning}: a rotor that transforms \emph{run} into \emph{running} encodes the specific morphological operation as a rotation in a semantically meaningful plane. This is a feature, not a limitation: the rotor is interpretable (its bivector identifies the plane of transformation), invertible ($\Rev{R}$ undoes the rotation), and composable with other rotors ($R_2 R_1$ is itself a rotor). The price of this structure is that rotors must either be specified by the theory (for qualia-based coercions) or learned from data (for contextual transformations) -- they do not come for free from random initialization. But the payoff is semantic transparency: one can inspect a learned rotor and determine \emph{what kind of transformation} it encodes.

\subsection{Unified Symbolic--Geometric Representation}

Unlike purely symbolic approaches or purely neural embeddings, GA supports continuous representations while preserving discrete algebraic structure. The geometric product is both a continuous operation (it maps continuous inputs to continuous outputs, and is differentiable) and an algebraically structured one (it respects grades, produces determinate results, and supports exact unbinding). What this creates, I believe, is a genuine bridge between formal semantics and distributional semantics -- not a compromise or approximation, but a single framework in which both symbolic operations (binding, application, quantification) and geometric operations (similarity, interpolation, gradient-based learning) coexist natively. This is the unification that the field has been working toward, and GA makes it algebraically precise.

\subsection{Broader Semantic Coverage}
\label{sec:broader}

The three mechanisms above -- wedge products for binding, contraction for application, rotors for transformation -- extend naturally across a wide range of semantic phenomena that standard embeddings handle poorly or not at all. Let me briefly sketch five such extensions to indicate the framework's scope; detailed formal treatments are developed in \citet{pustejovsky2026fga}.

\paragraph{Concepts as blades, not points.}
Standard distributional models represent a word as a single point in $\mathbb{R}^n$. This forces a commitment: \emph{dog} is one vector, regardless of whether we are thinking of the prototype (a Labrador), a borderline case (a chihuahua), or the category boundary (is a wolf a dog?). The psychological reality of graded category membership has been established since \citet{rosch1975}, and \citet{gardenfors2000} proposed modeling concepts as convex regions rather than points -- but without an algebraic framework for composition over regions.  \citet{widdows2004} took a related step within distributional semantics, using subspace projection and orthogonal complements to model word-sense distinctions and negation, anticipating part of the algebraic move that FGA generalizes to the full graded structure.  In FGA, a concept can be represented not as a point but as a \emph{blade} -- an oriented subspace of a given grade. A grade-$k$ blade $B = v_1 \wedge v_2 \wedge \cdots \wedge v_k$ defines a $k$-dimensional subspace of semantic space. Typicality becomes \emph{projection magnitude}: how much of an instance vector lies within the blade's subspace. Vagueness becomes \emph{partial containment}: a borderline case projects partially onto the blade. Conceptual hierarchy becomes \emph{blade inclusion}: the blade for \textsc{animal} is a subspace of the blade for \textsc{organism}, because it spans a subset of the relevant semantic dimensions. This replaces the single-point commitment of standard embeddings with a geometrically structured representation that natively supports graded membership, vagueness, and taxonomic hierarchy.

\paragraph{Artifacts as grade-2 bivectors: form bound to purpose.}
The grade structure encodes not just arity (how many arguments a predicate takes) but \emph{ontological complexity}. Natural kinds -- entities whose identity does not depend on a telic purpose -- are grade-1 vectors in the entity subspace: $\sem{\text{dog}} = \mathbf{d} \in V_E$. A dog is a dog whether or not anyone keeps it as a pet. The identity is intrinsic, carried entirely by the FORMAL quale, and the representation is a single vector. Artifacts are different. A knife is not merely a blade-shaped piece of metal; it is a blade-shaped piece of metal \emph{bound to the purpose of cutting}. Remove the purpose and you have scrap metal, not a knife. The artifact is constituted by the binding of physical form to telic function, and in FGA this constitution is the wedge product:
\begin{equation}
\sem{\text{knife}} = \mathbf{blade}_{\textsc{form}} \wedge \mathbf{cut}_{\textsc{telic}} \in V_E \wedge V_P, \quad \text{where } \mathbf{blade}_{\textsc{form}} \in V_E, \; \mathbf{cut}_{\textsc{telic}} \in V_P.
\end{equation}
The artifact is a grade-2 bivector -- intrinsically more complex than a natural kind, because it binds two kinds of information into a single structured object. The ``artifact subspace'' is not a primitive subspace of the base vector space $V$; it is a \emph{derived} subspace of $\Cl^2(V)$, specifically $V_E \wedge V_P$, constructed by the wedge product of two primitive subspaces. Artifacts \emph{emerge} at grade 2, just as artifacts in the world are constructed objects, not natural kinds. The qualia are recoverable by contraction: $\mathbf{blade}_{\textsc{form}} \lc \sem{\text{knife}} = \mathbf{cut}_{\textsc{telic}}$ (``What is its purpose?'') and $\mathbf{cut}_{\textsc{telic}} \lc \sem{\text{knife}} = \mathbf{blade}_{\textsc{form}}$ (``What is its form?''). The Generative Lexicon's qualia structure is thus realized as the subspace decomposition of the artifact bivector, with contraction as the quale-extraction operation.

\paragraph{Dot objects as mixed-grade multivectors.}
GL's dot objects -- entities that are simultaneously physical objects and information content, like \emph{book} -- receive a natural formalization as multivectors with support at \emph{multiple grades} and in multiple subspaces:
\begin{equation}
\sem{\text{book}} = \underbrace{\mathbf{b}_{\textsc{phys}}}_{\text{grade 1, } V_E} + \underbrace{\mathbf{b}_{\textsc{form}} \wedge \mathbf{b}_{\textsc{read}}}_{\text{grade 2, artifact}} + \underbrace{\mathbf{b}_{\textsc{info}}}_{\text{grade 1, } V_P}
\end{equation}
A book is simultaneously a physical object (grade 1 in $V_E$), an artifact constituted for reading (grade 2 in $V_E \wedge V_P$), and an information container (grade 1 in $V_P$). A verb like \emph{begin}, which selects for an event argument, triggers coercion by accessing the grade-2 artifact component and contracting to extract the telic event: contraction with the form vector recovers $\mathbf{b}_{\textsc{read}}$, the reading event. The mixed-grade representation means that the ``dual nature'' of a dot object is not an ambiguity to be resolved but a structural richness to be selectively accessed -- and the algebra itself, through grade projection and contraction, provides the selection mechanism.

\paragraph{Regular polysemy as rotor families.}
Many lexical alternations are not idiosyncratic but \emph{regular}: the animal/food alternation (\emph{chicken}, \emph{lamb}, \emph{turkey}), the container/contents alternation (\emph{bottle}, \emph{glass}, \emph{box}), the producer/product alternation (\emph{Samsung}, \emph{Honda}). The regularity of these patterns was first documented by \citet{apresjan1974} and formalized within Generative Lexicon theory \citep{pustejovsky1995} as systematic relations between qualia-derived senses. In FGA, each such alternation pattern is modeled as a \emph{rotor family} -- a set of rotors sharing a common rotation plane but potentially differing in angle. The animal/food alternation corresponds to a rotor $R_{\textsc{food}}$ that rotates within the plane spanned by the \textsc{living-entity} and \textsc{food-substance} directions. Applying this rotor to any animal vector produces the corresponding food reading: $R_{\textsc{food}}\, \sem{\text{chicken}}_{\textsc{animal}}\, \Rev{R}_{\textsc{food}} \approx \sem{\text{chicken}}_{\textsc{food}}$. The shared rotation plane captures the \emph{regularity} of the pattern, while the specific angle captures word-specific differences in how readily the alternation applies. Metaphor, by contrast, involves a \emph{large-angle} rotation through a plane that connects two otherwise distant semantic regions -- a continuum from literal polysemy (small rotation) through conventional metaphor to novel metaphor (large rotation), unified within a single algebraic mechanism.

\paragraph{Modification as geometric product.}
Adjective--noun modification is not a single operation: \emph{red car} (intersective), \emph{good knife} (subsective), and \emph{fake diamond} (privative) involve fundamentally different semantic relations \citep{kamp1995}. In FGA, these correspond to different components of the geometric product. Intersective modification uses the inner product component: $\langle \sem{\text{red}} \cdot \sem{\text{car}} \rangle$ projects the noun onto the property subspace, retaining only the region of semantic space where both noun and adjective apply. Subsective modification uses the full geometric product, where the adjective modulates the noun's qualia (a \emph{good knife} is good with respect to its telic quale -- cutting). Privative modification exploits the bivector component: \emph{fake diamond} introduces an oriented plane that rotates the noun away from its prototypical region while preserving its formal type. The three modification types, which require separate mechanisms in LA-based models, emerge as different grade projections of a single algebraic operation.

The signature choice introduced in Section~2.3 makes this algebraic distinction more precise. If entity-kind dimensions in $V_E$ have positive-squaring generators and telic-purpose dimensions in $V_P$ have negative-squaring generators, then the rotor that \emph{red} applies and the rotor that \emph{good} applies are \emph{algebraically different in kind}. The adjective \emph{red} operates within the property subspace $V_{\textsc{prop}}$ (positive-squaring), generating an \textbf{elliptic rotor} -- a smooth, cyclic rotation through color space. The adjective \emph{good}, applied to an artifact, operates in the $V_E \times V_P$ plane that spans one positive and one negative subspace, generating a \textbf{hyperbolic rotor} -- a non-cyclic boost that intensifies the alignment of physical form with telic purpose. A good knife is one whose form is more optimally aligned with the cutting function. The adjective \emph{fake} applies the same kind of hyperbolic rotor but with a negative parameter, \emph{reversing} the form-purpose alignment: a fake knife has the form of a knife but its telic binding is negated. The distinction between extensional modification (elliptic, cyclic, similarity-based) and intensional modification (hyperbolic, non-cyclic, function-oriented) is thus not merely a semantic label but an algebraic property determined by the signature of the subspaces the modifier connects.

\paragraph{Polar opposition as idempotent decomposition.}
Lexical semantics has long treated binary opposition as a foundational organizing relation \citep{lyons1977,cruse1986,murphy2003}: \emph{alive}/\emph{dead}, \emph{open}/\emph{closed}, \emph{true}/\emph{false}, and the asymmetrically lexicalized cases \emph{pregnant}, \emph{empty}, \emph{bald}.  The componential tradition \citep{katz1963,bierwisch1969,jackendoff1983} encoded the relation through binary features $[\pm F]$ in the lexical entry, treating each opposition pair as the two values of a privative feature.  Within formal semantics, \citet{kennedy2007} subsequently distinguished a class of \emph{absolute} gradable adjectives -- \emph{full}, \emph{empty}, \emph{open}, \emph{closed}, \emph{dead} -- whose scales are endpoint-closed and whose interpretations cluster at the scale endpoints; \citet{rotstein2004} and \citet{kennedy2005} refined this further into total versus partial classes, and \citet{horn2001} situated the same items within the broader logic of negation and polarity.  The intuition shared across these accounts is that polar predicates are not gradable predicates carved at a threshold but complementary, exhaustive, mutually exclusive partitions of an underlying state -- yet each account leaves open \emph{why} the scale is endpoint-closed, why complementarity holds, and what algebraic structure makes the predicate refuse the gradient interpretation.  The native algebraic answer is already present in $\Cl(V,Q)$: for any unit vector $e \in V_P$ with $e^2 = +1$, the algebra contains two complementary idempotents
\begin{equation}
  p_+ \;=\; \tfrac{1}{2}(1 + e), \qquad p_- \;=\; \tfrac{1}{2}(1 - e),
  \label{eq:polar-idem}
\end{equation}
satisfying $p_\pm^2 = p_\pm$, $\; p_+ + p_- = 1$, and $\; p_+\, p_- = 0$.  These project the algebra onto the $\pm 1$ eigenspaces of left-multiplication by $e$, partitioning entity-space exhaustively and exclusively.  A polar predicate is a mixed-grade idempotent rather than a grade-1 vector: $\sem{\text{alive}} = p_+$ and $\sem{\text{dead}} = p_-$ on a vital-state axis $e_v$.  The identification is precise: \citet{kennedy2007}'s absolute-gradable class is the class of predicates whose FGA realization is an idempotent on a positive-signature axis, and the endpoint-closed scale of the formal-semantic literature is the eigenspace structure $\pm 1$ of the algebraic encoding.  Asymmetric lexicalization -- English has \emph{pregnant} but no morpheme for the complementary pole, even though $p_+$ is sitting right there in the algebra -- corresponds to the markedness asymmetries documented since \citet{jakobson1971} and surveyed in \citet{battistella1996}: the lexicon names one of the two idempotents while the other remains algebraically present and recoverable.

Several lexical-semantic phenomena follow directly.  Complementarity and exhaustiveness become theorems from $p_+ + p_- = 1$ and $p_+ p_- = 0$, no longer stipulated by feature systems but proved from the algebra.  Sentential negation reduces to the swap $p_+ \leftrightarrow p_-$, equivalently multiplication by $-e$, giving \citet{horn2001}'s contradictory negation an algebraic realization rather than a logical primitive.  Aspectual transitions of the achievement class -- \emph{die}, \emph{arrive}, \emph{win} -- emerge as rotors that move an entity from one eigenspace to the other; this refines the BECOME analysis of \citet{dowty1979}, in which $\sem{\text{die}}(x) = \mathrm{BECOME}(\neg\sem{\text{alive}}(x))$, by giving BECOME explicit algebraic content as the rotor effecting the inter-eigenspace transition, and connects directly to the transition--process--state typology of \citet{pustejovsky1991}.  Privative modification (\emph{fake diamond}) becomes a flip of the FORMAL polar value while preserving the noun's other qualia, sharpening the privative subcase of the modification typology above.  Cruse's four-way typology of opposites \citep{cruse1986} -- complementaries, antonyms, converses, reversives -- correspondingly maps onto distinct algebraic shapes the algebra already supports: complementaries are idempotent pairs $p_\pm$; gradable antonyms are opposed grade-1 vectors $P$ and $-P$ on a shared axis; converses are role-permuted bivectors ($r_{\textsc{ag}} \wedge r_{\textsc{th}}$ vs.\ $r_{\textsc{th}} \wedge r_{\textsc{ag}}$); reversives are rotor inverses.  The four-way distinction, which requires separate stipulative machinery in feature-based and componential approaches, falls out of operations the algebra already supports.

\paragraph{The pseudoscalar and modal/discourse structure.}
The highest-grade element in $\Cl(V,Q)$ -- the \emph{pseudoscalar} $I = e_1 e_2 \cdots e_n$ -- has a natural semantic interpretation. It represents the \emph{orientation} of the entire semantic space: the global perspective from which all lower-grade meanings are assessed. In FGA, this can be exploited for modality and discourse structure. Possible worlds correspond to different pseudoscalar orientations: the actual world is one orientation, and accessible alternatives are related by reflections or rotations that reverse or tilt the pseudoscalar. Modal operators (``necessarily $p$,'' ``possibly $p$''), which in standard possible-worlds semantics \citep{kratzer1991} quantify over accessible worlds, act in FGA by testing a proposition's scalar value across a family of pseudoscalar orientations. At the discourse level, the pseudoscalar encodes the current discourse frame -- the shared orientation that determines how new sentences update the common ground \citep{stalnaker1978}. This extends FGA from sentence-level semantics into the domains of intensionality and discourse coherence, using the same algebraic machinery.

% ============================================================
\section{Functional Geometric Algebra: The Three-Layer Architecture}
\label{sec:architecture}

FGA extends geometric algebra toward a functional and type-sensitive semantics. The framework I propose here hypothesizes three operational layers, each exploiting a different aspect of the Clifford algebra:

\begin{enumerate}
\item \textbf{Layer 1: Truth-Conditional Semantics via Contraction.} Predicate application is modeled through the inner product and left contraction. Entities are grade-1 vectors in the entity subspace $V_E$; unary predicates are grade-1 vectors in $V_E \oplus V_P$, where the shared $V_E$ component enables contraction with entity arguments; binary relations are grade-2 bivectors in $V_E \oplus V_P$. Function application proceeds by contraction:
\begin{align}
\sem{\text{sleep}}(\sem{\text{john}}) &= \sem{\text{sleep}} \cdot \sem{\text{john}} \to \Cl^0 \quad \text{(scalar/truth value)}, \\
\sem{\text{loves}} \rc \sem{\text{mary}} &\to \Cl^1 \quad \text{(partial application: ``loves whom?'')}, \\
\sem{\text{john}} \lc (\sem{\text{loves}} \rc \sem{\text{mary}}) &\to \Cl^0 \quad \text{(full application)}.
\end{align}
The asymmetry of left contraction encodes argument order, addressing a persistent problem in tensor-based compositional semantics.

\item \textbf{Layer 2: Event Structure via the Wedge Product.} Thematic role binding uses the wedge product across a decomposed space $V_R \oplus V_E$, where $V_R$ contains role key vectors and $V_E$ contains entity vectors. Events are sums of role--entity bivectors, and fillers are recoverable by contraction (as shown in Section~5.4). This layer captures argument structure, aspectual composition, and event decomposition.

\item \textbf{Layer 3: Inner Application via Rotors.} As argued in Section~\ref{sec:oaia}, most natural language composition is not holistic but selective: predicates probe the internal structure of their arguments, composing with a specific quale, constitutive component, or functional dimension rather than with the argument as a whole. In FGA, this selective composition -- Inner Application (IA), as developed in the theory of Distributed Compositionality \citep{pustejovskyJezek2026} -- is realized as rotor action: a quale-targeting rotor transforms the argument's multivector before contraction applies. The qualia structure of Generative Lexicon theory \citep{pustejovsky1995,pustejovskyJezek2026} provides the inventory of available rotors: telic rotors (purpose-based access), agentive rotors (origin-based access), constitutive rotors (part/surface/material access), and formal rotors (taxonomic adjustment). Inner Application is the \emph{general} mode of composition; Outer Application (contraction alone, without a rotor) is the special case, restricted to predicates like ``exist'' or ``disappear'' that genuinely want the undifferentiated argument. When no quale-targeting rotor is active, the rotor is the identity, and the derivation reduces to Outer Application -- ensuring that FGA is conservative over standard compositional semantics.
\end{enumerate}

The three layers use three \emph{different} operations of the \emph{same} algebra. This, I believe, is the key architectural insight: GA does not require bolting on additional mechanisms for different semantic phenomena. Contraction, the wedge product, and the rotor sandwich are all native operations of $\Cl(V, Q)$, unified under the geometric product. The theoretical parsimony is considerable: a single algebraic substrate supports truth-conditional evaluation, event composition, and contextual modulation without requiring separate formalisms for each.

\paragraph{Relation to DisCoCat's tensor contraction.}
It is important to position FGA carefully with respect to DisCoCat \citep{coecke2010}, since that framework is the closest existing approach to what I am proposing. In DisCoCat, grammatical types are mapped via a monoidal functor $F: \mathbf{G} \to \mathbf{FdVect}$ to tensor product spaces, and composition proceeds by contracting matching indices via the epsilon maps ($\varepsilon: V^* \otimes V \to \mathbb{R}$) of a compact closed category. A transitive verb is represented as an order-3 tensor in $N \otimes S \otimes N$; applying it to a subject and object noun contracts two indices, yielding a vector in the sentence space $S$. This achieves the same type-theoretic effect as FGA's left contraction: applying a grade-1 argument to a grade-$k$ functor yields a grade-$(k{-}1)$ result, just as contracting a vector with an order-$k$ tensor yields an order-$(k{-}1)$ tensor. The parallel is structural and deep -- both implement function application as dimension reduction.

However, four differences are significant. First, DisCoCat requires an \emph{external} categorical apparatus -- the functor from grammar to vector spaces -- to determine which contractions to perform; in FGA the left contraction is intrinsic to the Clifford algebra and inherently asymmetric, natively encoding argument directionality without the left/right adjoint distinction ($n^l, n^r$) of pregroups. Second, DisCoCat's tensor product spaces grow combinatorially with arity (a transitive verb requires $n^3$ components), a well-known practical difficulty \citep{grefenstette2011}; FGA's graded representations remain bounded by the binomial coefficients $\binom{n}{k}$. Third, DisCoCat's compact closed category $\mathbf{FdVect}$ is symmetric monoidal, so the distinction between left and right arguments must be externally imposed by the grammar; Clifford left and right contractions ($\lc$ and $\rc$) are algebraically distinct. Fourth, handling phenomena beyond basic predicate--argument composition -- relative pronouns, coordination, quantification -- requires supplementary structures in DisCoCat (Frobenius algebras for copying and merging information \citep{kartsaklis2013,sadrzadeh2014}; bialgebras for quantifier semantics), whereas FGA's wedge products and rotors are native to the same Clifford algebra.

Coecke's later DisCoCirc framework \citep{coecke2021} moves considerably closer to FGA's spirit: word meanings become types whose states evolve, and sentences are gates in a circuit that update those states. This dynamic view parallels FGA's rotor-based state transformations, and I see the two programs as converging on similar insights from different mathematical starting points.

\begin{definition}[FGA Semantic Space]
An FGA semantic space is a tuple $(V, Q, V_E, V_P, V_R, V_S, V_C, \mathcal{R})$ where:
\begin{itemize}
\item $V = V_E \oplus V_P \oplus V_R \oplus V_S \oplus V_C$ is a vector space decomposed into entity, predicate-content, role, state, and control subspaces;
\item $Q$ is a quadratic form on $V$ determining the Clifford algebra $\Cl(V, Q)$;
\item $\mathcal{R} = \{R_{\textsc{telic}}, R_{\textsc{agent}}, R_{\textsc{const}}, R_{\textsc{formal}}, \ldots\}$ is a set of rotors implementing qualia-based semantic transformations.
\end{itemize}
\end{definition}

% ============================================================
\section{Compositional Sketches: Where LA Fails and FGA Succeeds}
\label{sec:sketches}

The preceding sections have developed FGA's formal machinery in the abstract. This section puts it to work on specific compositional phenomena -- chosen not for their exoticism but for their ordinariness. These are cases that any compositional semantics must handle, that LA-based models handle poorly or not at all, and that FGA addresses with operations already native to the algebra. The examples move from basic predicate--argument composition (Outer Application) through qualia-sensitive modification and complement coercion (Inner Application, cross-subspace) to selective verb--argument composition (Inner Application, intra-subspace), illustrating all three layers of the FGA architecture and demonstrating that Inner Application is the general mode of composition, not a special-case mechanism.

\subsection{Verb--argument composition: the baseline}

I begin with the simplest case to establish that FGA handles basic composition cleanly before moving to cases where it offers genuine advantages. Consider \emph{John loves Mary}, analyzed in Section~5. The lexical inventory assigns entity vectors $j, m \in \Cl^1(V_E)$, a predicate bivector $\ell \in \Cl^2(V_E \oplus V_P)$, and role keys $r_{\textsc{ag}}, r_{\textsc{th}} \in \Cl^1(V_R)$. The two compositional outputs are:

\paragraph{Truth-conditional meaning} (Layer 1, via contraction):
\begin{equation}
\sem{\text{John loves Mary}}_t = j \lc (\ell \rc m) \in \Cl^0.
\end{equation}
Right contraction saturates the object; left contraction saturates the subject. The asymmetry of the two contractions encodes argument order without any external categorical apparatus.

\paragraph{Event structure} (Layer 2, via wedge product):
\begin{equation}
\sem{\text{John loves Mary}}_e = (r_{\textsc{ag}} \wedge j) + (r_{\textsc{th}} \wedge m) \in \Cl^2(V).
\end{equation}
Each binding is an oriented bivector; fillers are exactly recoverable by contraction.

Even at this basic level, FGA provides what LA cannot: native encoding of argument order (left vs.\ right contraction), exact role--filler binding (wedge product with exact unbinding via contraction), and type checking (grade arithmetic and subspace compatibility, as developed in Section~5.8, ensure that only well-typed compositions yield nonzero results). In LA, all of these must be learned or externally imposed.

\subsection{Qualia-sensitive modification: \emph{good} $+$ N}

Now consider a case where LA fails conspicuously. The adjective \emph{good} is subsective: a \emph{good knife} is good \emph{as a knife}, not good in some absolute sense \citep{kamp1995}. More precisely, \emph{good} evaluates its argument relative to a contextually selected dimension of quality -- and the selection is governed by the noun's qualia structure \citep{pustejovsky1995,pustejovskyJezek2026}:

\begin{quote}
\emph{good pencil} $\approx$ ``writes well'' \quad (telic quale: what it is for)\\
\emph{good book} $\approx$ ``reads well'' or ``well written'' \quad (telic or agentive quale)\\
\emph{good sandwich} $\approx$ ``tastes good'' or ``well made'' \quad (telic or agentive quale)
\end{quote}

\paragraph{The LA problem.}
In a standard embedding model, \emph{good} is a single vector $\mathbf{g} \in \mathbb{R}^d$. Composition with a noun is typically additive ($\mathbf{g} + \mathbf{n}$) or multiplicative ($\mathbf{g} \odot \mathbf{n}$). Neither mechanism provides a principled way to rotate the adjective's evaluation dimension based on the noun's internal structure. The model must learn, from distributional data alone, that \emph{good pencil} activates writing-related features while \emph{good sandwich} activates eating-related features.

The most sophisticated LA-based response to this problem comes from the Distributional Functional Model (DFM) tradition \citep{baroni2014}, which represents predicates not as vectors but as matrices or higher-order tensors. In this framework, an adjective like \emph{good} is a matrix $G \in \mathbb{R}^{d \times d}$ that operates on the noun vector by matrix-vector multiplication: $\sem{\text{good pencil}} = G\, \mathbf{n}_{\textsc{pencil}}$. This is a genuine advance over additive models -- it captures the asymmetry between functor and argument -- and it parallels function application in simple type theory, as LS observe in their comprehensive treatment of distributional semantics \citep{lenci2023}. But the matrix $G$ is a single learned linear map. It applies the \emph{same transformation} to every noun, regardless of the noun's internal semantic structure. The matrix has no mechanism for selecting which dimension of quality is relevant: it cannot ``know'' that \emph{pencil} should activate writing-quality while \emph{sandwich} should activate eating-quality, unless these distinctions have been captured in the noun vectors themselves during training. If the training data is sparse for a particular combination, the model has no structural fallback.

\paragraph{The FGA account.}
In FGA, \emph{good} is a grade-1 predicate vector $g \in \Cl^1(V_P)$, encoding a general evaluative direction. Each noun carries qualia rotors as part of its lexical entry -- these are the $\mathcal{R}$ component of the FGA semantic space defined in Section~\ref{sec:architecture}. The telic rotor of \emph{pencil}, $R_{\textsc{tel}}^{\textsc{pencil}}$, encodes the pencil's characteristic function (writing). Subsective modification is then a rotor sandwich:
\begin{equation}
\sem{\text{good pencil}} = R_{\textsc{tel}}^{\textsc{pencil}}\, g\, \Rev{R}_{\textsc{tel}}^{\textsc{pencil}} \in \Cl^1(V_P).
\end{equation}
The telic rotor rotates the evaluative direction of \emph{good} toward the writing-quality region of predicate space. For \emph{good sandwich}, a different telic rotor $R_{\textsc{tel}}^{\textsc{sandwich}}$ rotates \emph{good} toward the eating-quality region. The result is grade-preserving (the modified meaning is still a predicate), norm-preserving (the ``strength'' of the evaluation is not distorted), and invertible (the original sense of \emph{good} is recoverable). Three different nouns yield three different orientations of the same adjective -- not by learning three separate interactions, but by the algebraic action of three different qualia rotors on a single lexical entry.

The case of \emph{good book}, where both the telic (``reads well'') and agentive (``well written'') readings are available, is handled by the composition of rotors: $R_{\textsc{ag}}^{\textsc{book}} \circ R_{\textsc{tel}}^{\textsc{book}}$ yields different final orientations depending on which rotor is contextually selected or weighted -- a natural account of the polysemy of the modified expression.

This qualia-driven account of subsective modification speaks directly to the distinction \citet{mcnally2017} 
draw between lexically licensed patterns of composition and more referent-sensitive, contextually induced construals. On the present view, the default reading of \emph{good knife} is anchored in the noun's telic organization, which is encoded in the lexical entry and therefore available compositionally by default. A more context-dependent interpretation---for example, one that evaluates visual sharpness rather than cutting function---emerges when discourse conditions favor a different qualia-guided reorientation. 
 The OA/IA distinction developed in \citet{pustejovskyJezek2026} makes this contrast precise: OA yields the default lexically grounded composition, while IA provides the pragmatically licensed enrichment.
 
\subsection{Complement coercion: \emph{open} $+$ NP}

Finally, consider a case that exercises all three layers simultaneously. The verb \emph{open} combines with different objects to yield strikingly different event types:

\begin{quote}
\emph{open the door} $\approx$ physical state change \quad (formal quale: door has open/closed states)\\
\emph{open a book} $\approx$ begin reading \quad (telic quale: what a book is for)\\
\emph{open a trail} $\approx$ create by clearing \quad (agentive quale: how a trail comes into being)\\
\emph{open a window} $\approx$ physical change $+$ enable ventilation \quad (formal $+$ telic)
\end{quote}

\paragraph{The LA problem.}
In LA, \emph{open} has one vector, and these different event types must be inferred entirely from collocational patterns. There is no mechanism in the representation itself that distinguishes a state-change reading from a creation reading from an inceptive reading. The DFM approach improves on this by representing a transitive verb as a third-order tensor $\mathcal{T} \in \mathbb{R}^{d \times d \times d}$ that composes with subject and object vectors \citep{baroni2014}. This captures argument structure -- \emph{dogs chase cats} produces a different result from \emph{cats chase dogs} -- but at the cost of $d^3$ parameters per verb, an exponential increase in dimensionality that has been a persistent practical obstacle \citep{grefenstette2011}. More critically, the tensor is a fixed learned object. It applies the same bilinear map regardless of which noun fills the object position. The DFM framework provides no mechanism for the complement's internal lexical structure to modulate the verb's event type: it cannot represent the fact that \emph{open a book} accesses the book's telic quale while \emph{open a trail} accesses the trail's agentive quale, because the nouns have no qualia structure for the verb to interact with. The model must learn each verb--object combination as a separate distributional pattern.

\paragraph{The FGA account.}
In FGA, \emph{open} is a grade-2 bivector $\omega \in \Cl^2(V_E \oplus V_P)$ -- a binary relation whose core meaning is something like ``make accessible.'' Each object noun carries qualia rotors encoding its characteristic properties, and the compositional outcome depends on which quale the verb accesses:

\emph{Open the door.} The door's formal quale includes an open/closed state opposition. No coercion is needed: the verb's core meaning applies directly via contraction, and the event structure is a straightforward state change.
\begin{equation}
\sem{\text{open the door}}_e = (r_{\textsc{ag}} \wedge j) + (r_{\textsc{th}} \wedge d) + (r_{\textsc{res}} \wedge \omega_{\textsc{state}}).
\end{equation}

\emph{Open a book.} The book's formal quale does not include an open/closed state opposition in the relevant sense. Instead, its telic quale (reading) is accessed, and the telic rotor $R_{\textsc{tel}}^{\textsc{book}}$ acts on the verb's event representation, rotating it from a physical-state-change reading into an inceptive reading -- ``begin the activity the book is for'':
\begin{equation}
\omega' = R_{\textsc{tel}}^{\textsc{book}}\, \omega\, \Rev{R}_{\textsc{tel}}^{\textsc{book}}.
\end{equation}
The rotated $\omega'$ now encodes a begin-reading event. 
This is complement coercion in the GL sense \citep{pustejovsky1995}: the object's qualia structure transforms the verb's event type. In the Distributed Compositionality framework of \citet{pustejovskyJezek2026}, this is an instance of Inner Application -- a licensed, optional enrichment in which the argument's qualia content resolves underspecification in the verbal predication.

\emph{Open a trail.} The trail's agentive quale (how it comes into being: by clearing, blazing) is accessed, and the agentive rotor $R_{\textsc{ag}}^{\textsc{trail}}$ rotates the verb into a creation reading:
\begin{equation}
\omega'' = R_{\textsc{ag}}^{\textsc{trail}}\, \omega\, \Rev{R}_{\textsc{ag}}^{\textsc{trail}}.
\end{equation}
The same verb, the same algebraic operation, but a different rotor -- and the result is a different event type, compositionally derived from the object's lexical semantics.

\subsection{Inner Application as the general mode: ``wipe the table'' and ``tie your shoe''}

The preceding examples -- qualia-sensitive modification, complement coercion -- are standardly treated as special compositional phenomena requiring dedicated mechanisms. What FGA reveals, through the Outer/Inner Application distinction developed in Section~\ref{sec:oaia}, is that these are instances of the \emph{same} compositional mechanism -- Inner Application -- that drives the vast majority of verb-argument interaction. The present section demonstrates this with two ordinary transitive verbs that are not normally treated as involving coercion but that exhibit precisely the same selective structure.

Consider ``wipe the table.'' The table is a complex physical object with constitutive structure: bulk geometry, legs, a surface, a material composition. ``Wipe'' does not predicate an action on the table as an undifferentiated whole -- it targets the table's \emph{surface}. This is Inner Application with an intra-subspace rotor: the verb carries a constitutive rotor $R_{\textsc{const}}^{\textsc{wipe}}$ that acts within $V_E$, foregrounding the surface-component dimensions of the argument's entity vector:
\begin{equation}
R_{\textsc{const}}^{\textsc{wipe}} \; \mathbf{t} \; \Rev{R}_{\textsc{const}}^{\textsc{wipe}}
\end{equation}
The result is still a grade-1 entity vector -- the rotor is grade-preserving -- but its orientation within $V_E$ has shifted: the surface features are now compositionally active. Contraction then saturates the argument position with this rotated representation. The same rotor applies in ``wipe your face'': it projects the surface of whatever object it composes with. What varies across arguments is not the verb's operation but the constitutive structure the rotor encounters.

Now consider ``tie your shoe.'' Here Inner Application targets a more specific constitutive component: a subpart that meets geometric and physical conditions for being tie-able -- flexible, elongated, rope-like. The verb carries a constitutive rotor $R_{\textsc{const}}^{\textsc{tie}}$ that probes the argument's constitutive structure for a component meeting these conditions:
\begin{equation}
R_{\textsc{const}}^{\textsc{tie}} \; \mathbf{s} \; \Rev{R}_{\textsc{const}}^{\textsc{tie}}
\end{equation}
If the shoe's multivector has a constitutive component with substantial support along the flexibility and elongation dimensions -- the laces -- then the rotated representation has non-trivial magnitude along the verb's selectional axis, and composition succeeds. If the argument lacks such a component, as with ``tie the brick,'' the rotated representation has near-zero support, and contraction yields a near-zero result. The anomaly falls out of the algebra as a geometric mismatch, not a stipulated constraint violation.

The pattern generalizes across the lexicon. ``Drink the glass'' is Inner Application targeting the glass's contents (constitutive subselection). ``Read the wall'' targets inscriptions on the wall's surface (constitutive projection followed by informational-content access). ``Bake the cake'' accesses the cake's Agentive quale -- the creation process. In each case, the verb carries a quale-targeting rotor that foregrounds the relevant component; contraction then saturates the argument position with the result. Selection in FGA is not a filter but a \emph{geometric operation}: the verb actively probes the argument's internal structure through its rotor, and compositional success depends on geometric compatibility between the rotor's orientation and the argument's subspace support.

\subsection{What these examples show}

The four cases are not four separate mechanisms. They are instances of a single compositional architecture, distinguished by which mode of application they employ and which geometric target the rotor selects:

\begin{itemize}
\item \textbf{Verb--argument composition} (``John loves Mary'') is \emph{Outer Application}: contraction alone (Layer 1) with wedge-product binding (Layer 2). The predicate takes its arguments as undifferentiated wholes. This is the baseline -- the special case where no rotor is needed, and FGA reduces to classical function application with the added benefits of native argument-order encoding and exact role--filler recovery.

\item \textbf{Selective verb--argument composition} (``wipe the table,'' ``tie your shoe'') is \emph{Inner Application, intra-subspace}: a constitutive rotor acts within $V_E$ to foreground a specific component of the argument (surface, subpart, contents) before contraction applies. This is the general mode of verb-argument interaction -- most predicates probe the internal structure of their arguments rather than taking them whole.

\item \textbf{Subsective modification} (``good knife'') is \emph{Inner Application, cross-subspace}: the noun's Telic rotor acts from $V_E$ to $V_P$, rotating the adjective's evaluative direction toward the quale-appropriate dimension of quality. This is where FGA surpasses LA, which has no mechanism for principled, structured, qualia-driven reorientation of meaning.

\item \textbf{Complement coercion} (``open a book,'' ``begin the book'') is \emph{Inner Application, cross-subspace}, exercising all three layers: contraction for argument saturation, wedge products for event-structural binding, and a qualia rotor for event-type shifting. This is where FGA surpasses not only LA but also most existing compositional frameworks, which require separate coercion modules external to the core algebra.
\end{itemize}

The organizing principle is the Outer/Inner Application distinction developed in Section~\ref{sec:oaia}. What the examples demonstrate is that Inner Application -- the rotor-then-contraction sequence -- is not a special mechanism activated by type mismatch. It is the \emph{default} mode of composition. Outer Application is the exception, restricted to the small class of predicates that genuinely want their arguments as undifferentiated wholes. Coercion, subsective modification, and constitutive projection are all species of the same mechanism, differing only in the geometric target of the rotor (intra-subspace vs.\ cross-subspace) and the quale accessed (Telic, Constitutive, Agentive, Formal).

The examples are deliberately ordinary -- these are phenomena that occur in every text, not exotic corner cases. That ordinariness is the point. If FGA's algebraic structure is needed to handle \emph{good pencil}, \emph{wipe the table}, and \emph{open a book} -- compositions that any competent speaker processes effortlessly and that any adequate semantics must represent -- then the argument for moving beyond LA is not a theoretical luxury but a practical necessity.

% ============================================================
\section{GA and Modern Neural Architectures}

A natural concern is whether GA-based semantics can integrate with existing neural systems. The answer, I want to suggest, is more encouraging than one might expect: key GA operations are \emph{already implicit} in current transformer architectures.

\subsection{Rotary Position Embeddings as Rotors}

Rotary Position Embeddings (RoPE) \citep{su2024}, now standard in models such as LLaMA, PaLM, and GPT-NeoX, encode token positions as rotations applied to query and key vectors. For a $d$-dimensional embedding, RoPE pairs adjacent dimensions $(2i, 2i+1)$ and applies a 2D rotation:
\begin{equation}
\begin{pmatrix} q_{2i}' \\ q_{2i+1}' \end{pmatrix} = \begin{pmatrix} \cos m\theta_i & -\sin m\theta_i \\ \sin m\theta_i & \cos m\theta_i \end{pmatrix} \begin{pmatrix} q_{2i} \\ q_{2i+1} \end{pmatrix},
\end{equation}
where $m$ is the position index and $\theta_i$ is a frequency parameter.

This is precisely a rotor operation in $\Cl(2, 0, 0)$ applied independently across dimension pairs. The connection between RoPE and Clifford rotors has been made explicit by recent work: \citet{care2025} show that RoPE embeddings are restricted rotors in $\Cl(2,0,0)$ and generalize them to Clifford Algebraic Rotary Embeddings (CARE) in $\Cl(3,0,0)$ and beyond, encoding positional information across multiple grades and arbitrary dimension triples. The core insight of RoPE -- that relative position information emerges from rotor composition ($R_m^{-1} R_n = R_{n-m}$) -- is a native property of the Clifford algebra. RoPE can be understood as exploiting GA structure, but in a restricted way: rotations are applied only within fixed 2D planes, with no interaction between planes. A full GA treatment would allow rotations in arbitrary planes of the semantic space, potentially encoding not just positional but semantic relationships through the same mechanism, as CARE begins to explore.

\subsection{Attention as Geometric Interaction}

Multi-head attention projects inputs into multiple subspaces and computes interactions within each. This admits a suggestive GA \emph{analogy}: each attention head operates within a different learned subspace, and the multi-head mechanism combines information from subspaces of different orientations -- structurally parallel to how a Clifford algebra decomposes into blades of different grades and orientations. While attention heads project into ordinary linear subspaces rather than blades of a Clifford algebra, the analogy suggests that reformulating attention in GA terms could make the geometric structure of multi-head attention explicit and exploitable.

The attention score $\text{softmax}(QK^\top / \sqrt{d_k})$ computes a scaled inner product -- the grade-0 (scalar) part of the geometric product. The information discarded by this projection is precisely the grade-2 (bivector) component that encodes relational orientation. Incorporating the full geometric product into attention would allow heads to capture not just ``how similar'' two tokens are, but ``in what oriented direction'' they relate.

This observation has received increasing empirical support from several recent architectures that move toward the full geometric product. The Geometric Algebra Transformer (GATr; \citet{brehmer2023}) develops a transformer architecture directly in projective geometric algebra, so that the model's internal states and interactions are expressed in multivector form rather than in ordinary vector space alone. This provides strong evidence that Clifford-algebraic structure can be incorporated into transformer design in a way that remains practically scalable.
 More recently, \citet{hirst2026} propose Versor, a sequence architecture formulated in conformal geometric algebra, in which temporal state updates are carried out through rotor-based transformations. Their design is especially relevant here because it treats rotor evolution and geometric-product interaction as first-class computational primitives for sequential modeling.
In the experiments reported by the authors, Versor outperforms several transformer, graph, and GA-based baselines across a range of benchmarks, while doing so with markedly smaller parameter counts than conventional transformer models. CliffordNet \citep{ji2026}, a vision architecture centered on the Clifford geometric product, provides complementary evidence from a neighboring domain. By making both the scalar and bivector components of feature interaction computationally explicit, it reports a favorable efficiency--performance balance and suggests that the antisymmetric relational component of the representation carries substantial discriminative value in its own right.
 For NLP, the collective implication of these results is direct: the bivector component that current attention mechanisms discard is precisely the relational orientation information that compositional semantics requires.

\subsection{Multi-head attention and mixture-of-experts as implicit subspace selection}

A further connection deserves attention, because it suggests that existing transformer architectures may already be performing something like grade- or subspace-sensitive processing, albeit implicitly. The extensive ``BERTology'' literature \citep{rogers2021} has documented that different attention heads in transformer models specialize for different linguistic functions: some heads attend to syntactic dependencies, others to coreference, others to semantic role structure, and still others to positional or discourse-level features. These specializations are not architecturally imposed -- they emerge from learning. But they are structurally parallel to FGA's subspace decomposition: each head operates within a learned projection subspace, and the multi-head mechanism combines information from subspaces of different orientations, much as a Clifford algebra decomposes into blades of different grades and orientations.

Similarly, mixture-of-experts (MoE) systems route tokens to different expert networks based on learned gating functions. The gating decision is, in effect, a form of contextual subspace selection: different experts activate different ``factors'' of word meaning depending on context. Recent work has shown that this gating can capture fine-grained sense distinctions -- selecting different expert pathways for ``Romeo'' in ``Romeo and Juliet'' versus ``Alfa Romeo,'' for instance \citep{widdows2025}. In FGA terms, these expert pathways correspond to different qualia-like projections of the word's multivector representation: the literary context activates the character dimensions while the automotive context activates the brand dimensions.

The connection suggests a clear experimental question that I think is worth pursuing: can we set up a comparison between rotor-based and projection-based mechanisms for selecting the contextually appropriate aspects of a word's meaning? In the current transformer paradigm, contextual modulation operates through learned linear projections (the $Q$, $K$, $V$ matrices). In FGA, the corresponding operation is a rotor sandwich -- a grade-preserving, invertible transformation. The two mechanisms make different predictions about what is preserved under contextual modulation (rotors preserve norm and grade; projections do not) and about recoverability (rotor transformations are invertible; projections are not). An experiment that trains both mechanisms on sense disambiguation or qualia-sensitive modification tasks, and compares their compositional generalization, would provide direct evidence for whether the richer algebraic structure of rotors offers advantages beyond what linear projections achieve.

\subsection{Existing GA-Neural Architectures}

Recent work has demonstrated the feasibility of GA operations within neural networks:

\begin{itemize}
\item \citet{brandstetter2023} introduce Clifford neural layers -- Clifford convolutions and Clifford Fourier transforms operating on multivector-valued feature maps -- demonstrating that the geometric product can serve as a trainable primitive in deep networks for PDE modeling. \citet{ruhe2023} extend this line with Clifford Group Equivariant Neural Networks, providing equivariance guarantees and grade-based activation functions. Together, these works establish the computational feasibility and differentiability of GA operations within standard neural training pipelines.

\item \citet{brehmer2023} propose the Geometric Algebra Transformer (GATr), a transformer architecture formulated in projective geometric algebra $(G_{3,0,1})$, with grade-sensitive normalization, equivariant attention, and multivector-valued interactions. Taken together, these design choices show that transformer computation can be carried out directly over Clifford-algebraic representations without abandoning scalability. \citet{dehaan2024} systematically compare Euclidean, projective, and conformal algebras within this architecture, finding nontrivial tradeoffs in expressivity and efficiency -- empirical evidence directly relevant to the question of signature selection raised in Section~\ref{sec:implementation}.

\item \citet{white2024} suggests replacing standard vector embeddings in transformers with \emph{spinor embeddings} in Clifford algebras, using rotors for positional encoding and geometric inner products for attention computation. This is the closest existing work to an FGA-transformer architecture.

\item In knowledge graph embeddings, \citet{xu2020} introduced GeomE, using multivector representations (scalars, vectors, bivectors, trivectors) with the geometric product for scoring relational triples, demonstrating that GA natively captures symmetry, inversion, and composition patterns. \citet{demir2023} further showed that the Clifford algebra signature $(p,q)$ itself can be learned as part of the embedding model, parameterizing the algebra to fit the relational structure of the data.

\item \citet{ji2026} introduces CliffordNet, a vision backbone organized around the Clifford geometric product as its primary interaction operator, rather than the usual combination of attention and large feed-forward expansion blocks. By separating the geometric product into scalar and bivector contributions and computing these efficiently, the model achieves linear complexity $O(N)$ while preserving both the symmetric and antisymmetric structure of feature interaction. The reported results indicate a notably strong accuracy--efficiency tradeoff in vision, and the broader lesson for NLP is that geometric-product interaction may absorb functions that are ordinarily distributed across separate spatial and channel-mixing components in transformer architectures.

\item Clifford algebra libraries (e.g., \texttt{clifford}, \texttt{geometric-algebra-attention}) now support GPU-accelerated geometric product computation, making large-scale implementation practical.
\end{itemize}

The convergence here is, I think, no longer merely suggestive -- it is empirical. The ML community is independently moving toward geometric algebra from the engineering side: GATr \citep{brehmer2023} shows that an entire transformer can operate natively on multivectors; Versor \citep{hirst2026} demonstrates that rotor-based state evolution achieves dramatic parameter efficiency and zero-shot generalization; RoPE and CARE exploit restricted and generalized rotors for positional encoding; CliffordNet \citep{ji2026} further suggests that geometric-product-based interaction may absorb functions that are ordinarily distributed across separate attention and feed-forward components; and Clifford neural layers \citep{brandstetter2023} and equivariant networks \citep{ruhe2023} provide the differentiable infrastructure. FGA approaches the same territory from the linguistic and semantic side, providing the compositional and type-theoretic motivations that these engineering advances lack. The opportunity for a principled synthesis -- where the algebraic structure that makes GATr scalable and Versor efficient is the same structure that makes FGA compositional -- is immediate.

% ============================================================
\section{Comparison with Alternative Frameworks}

Several frameworks have attempted to bridge formal and distributional semantics. It is important to position FGA clearly in relation to these, both to acknowledge the considerable debts it owes and to identify what it uniquely provides.

\begin{center}
\renewcommand{\arraystretch}{1.3}
\begin{tabular}{@{}p{2.2cm}p{2.0cm}p{2.0cm}p{2.0cm}p{2.0cm}p{2.0cm}@{}}
\toprule
& \textbf{Montague} & \textbf{DisCoCat} & \textbf{TPR} & \textbf{Concept.\ Spaces} & \textbf{FGA} \\
\midrule
\textbf{Represent.} & Functions/ relations & Tensors in $\mathbf{FdVect}$ & Tensor products & Regions in metric spaces & Multivectors in $\Cl(V, Q)$ \\
\textbf{Composit.} & Function applic. & Tensor contraction & Outer product + unbind & Intersection/ blending & Geom.\ product (contract., wedge, rotor) \\
\textbf{Type system} & Lambda calculus & Gramm.\ types (Lambek) & Role--filler types & Partially formalized & Grade structure + qualia \\
\textbf{Coercion} & No native mech. & No native mech. & No native mech. & Not formalized & Rotors \\
\textbf{Learnab.} & Not applicable & Partial & Yes & Partial & Yes (differentiable) \\
\textbf{Event str.} & External (Davidson) & Not native & Not native & Partial \citep{gardenfors2012} & Native (wedge product) \\
\textbf{Invertib.} & Yes (lambda) & Partial & Approximate & No & Yes (rotor inverse) \\
\textbf{Scalability} & Intractable & $O(n^k)$ tensors & $O(n^k)$ tensors & Good & $O(\binom{n}{k})$ grades \\
\bottomrule
\end{tabular}
\end{center}

\paragraph{DisCoCat.} The categorical compositional distributional (DisCoCat) framework of \citet{coecke2010} is, as I have noted, the closest existing approach to FGA. Both use algebraic operations over vector spaces for compositional semantics, and both exploit structured multi-linear composition guided by grammatical type. The deep structural parallel between DisCoCat's tensor contraction and FGA's graded contraction is discussed in Section~\ref{sec:architecture}. The key differences bear repeating here: (i) FGA's contraction is intrinsic and inherently asymmetric, while DisCoCat requires external categorical structure (the monoidal functor from grammar); (ii) FGA provides native type coercion (rotors) and event structure (wedge products), which DisCoCat lacks and must supplement with Frobenius algebras and other categorical apparatus \citep{kartsaklis2013,coecke2017}; (iii) FGA's grade-bounded representations avoid the combinatorial tensor-order explosion that is a persistent practical challenge for DisCoCat \citep{grefenstette2011}; and (iv) all FGA operations live in a single Clifford algebra, whereas DisCoCat distributes meaning across a family of tensor product spaces connected by categorical morphisms. The later DisCoCirc framework \citep{coecke2021}, in which sentences are gates that update word states, converges with FGA's rotor-based dynamics.

I should note, in fairness, that the practical case for any explicit compositional framework -- DisCoCat, FGA, or otherwise -- faces a genuine challenge from the empirical success of large language models. LLMs are currently the best parsers, the best semantic role labelers, and the best at compositional generalization on standard benchmarks, despite having no explicit type system, no named compositional operations, and no grade structure. If we ask an LLM to analyze the grammatical structure of a sentence, it can do so -- unlike most untrained humans -- but it clearly does not \emph{need} to in order to process the sentence, which is also, interestingly, like most untrained humans. This raises a legitimate question: does FGA's type-graded compositional architecture provide something that LLMs genuinely lack, or is it an explanatory formalism that describes structure the LLM has already learned implicitly?

My answer is twofold. First, the question ``does it work in practice?'' and the question ``does it provide the right theoretical framework?'' are different questions, and both matter. LLMs work in practice by brute-force induction over astronomical quantities of data. FGA proposes to provide the \emph{algebraic inductive bias} that would allow smaller models to achieve compositional generalization with less data -- not to replace LLMs but to identify the structural principles they are approximating. Second, and more specifically: the phenomena that motivate FGA -- type coercion, qualia-sensitive modification, invertible meaning modulation, exact role-filler binding -- are precisely the phenomena where LLMs are known to be most brittle. Compositional generalization benchmarks (COGS, SCAN, gSCAN) consistently reveal that LLMs struggle with systematic compositionality, and the failures are not random -- they cluster around the kinds of type-shifting and role-structural operations that FGA handles natively. The case for FGA is not that LLMs cannot process language; it is that the algebraic structure FGA makes explicit is the structure that compositional generalization requires, and that providing it as an inductive bias should improve generalization where flat vector spaces fail.

\paragraph{Type Composition Logic.} Asher's Type Composition Logic \citep{asher2011}, developed  in close dialogue with Generative Lexicon theory, provides a sophisticated type-theoretic treatment of coercion, co-composition, and selectional restrictions. In TCL, type mismatches trigger type presuppositions that are resolved through accommodation — a logical operation that shifts the argument into the type required by the functor. This shares with FGA the fundamental insight that composition is not always simple function application but sometimes requires type adjustment. The difference is in the mechanism: TCL's accommodation is a logical operation over symbolic type structures, external to the representational substrate of the meanings being composed. In FGA, the corresponding operation — rotor action — is internal to the Clifford algebra, the same algebraic framework that handles similarity, binding, and application. The geometric product does not require a separate accommodation module; coercion is a native algebraic operation. More recently, \citet{asher2016} have explored integrating distributional vectors into TCL to represent what they call ``internal content,'' recognizing that the purely symbolic type structures of TCL do not capture the graded, usage-sensitive aspects of lexical meaning. FGA addresses this integration from the opposite direction: rather than adding distributional vectors to a symbolic type system, it provides an algebraic substrate — the Clifford algebra — in which both symbolic structure (grade, subspace, type) and continuous geometric structure (similarity, orientation, rotation) are native.

\paragraph{Combinatory Categorial Grammar.} Steedman's CCG \citep{steedman2000} provides a type-driven compositional framework in which a small set of combinatory rules (application, composition, type-raising) operates over syntactic categories that are themselves type specifications. The parallel with FGA is instructive: in CCG, the category $S\backslash NP$ specifies that a verb combines with an NP to its left to yield a sentence, and the directionality of the slash encodes argument order. In FGA, the corresponding operation is contraction, and argument directionality is encoded by the distinction between left and right contraction ($a \lc B$ vs.\ $B \rc a$). CCG's type-raising, which transforms an entity-type NP into a functor that takes a predicate as argument, has a natural analogue in FGA's grade-shifting operations. Where the frameworks diverge is in the representational substrate: CCG's categories are symbolic objects with no native geometric content -- they encode combinatory potential but not similarity, gradedness, or continuous semantic structure. FGA embeds the same combinatory logic within a continuous geometric algebra, so that type-driven composition and distributional similarity coexist in a single representation.

\paragraph{Tensor Product Representations.} Smolensky's TPRs \citep{smolensky1990} are the most direct precursor to FGA's binding mechanism, and the intellectual debt should be acknowledged. Both use a product operation on role and filler vectors to create structured representations. The critical difference is algebraic: TPRs use the tensor (outer) product, which increases tensor order with each binding and requires approximate unbinding in the compressed case; FGA uses the wedge product, which remains within the bounded grade structure of the Clifford algebra ($\binom{n}{k}$ components for grade-$k$ elements) and supports exact recovery via contraction. TPRs also lack native operations for type coercion or systematic transformation -- limitations that FGA's rotor mechanism addresses.

\paragraph{TTR--SPA hybrid approaches.} A particularly instructive comparison is with the recent work of \citet{larsson2025}, who bridge Cooper's Type Theory with Records (TTR) and Eliasmith's Semantic Pointer Architecture (SPA) to model question answering within a neurally grounded vector symbolic architecture. Their approach shares FGA's central ambition: to combine typed functional structures with distributed vector representations in a single framework. TTR record types -- labelled sets of type judgments -- are mapped onto superpositions of role--filler convolutions in HRR, and a lambda-calculus approximation is constructed from the same vector operations. The convergence in goals is striking, and the TTR--SPA program demonstrates that the demand for typed compositional structure within distributional representations is widely felt, not an idiosyncrasy of the FGA proposal. The VSA program's foundational insight --  that distributed representations require a genuine algebra over vectors, not just a vector space -- is one FGA fully inherits; the argument here concerns which algebra best realizes that program, not whether such an algebra is needed.
Where the approaches diverge is in the algebraic substrate. Because the SPA relies on HRR's circular convolution, binding is commutative (losing argument order), unbinding is approximate (requiring cleanup memories), and functional application must be simulated through slot-filling operations rather than arising natively from the algebra. FGA's Clifford algebra provides what the TTR--SPA approach must work around: antisymmetric binding (preserving argument order), exact unbinding (via contraction), native function application (via grade reduction), and type coercion (via rotors). I see the TTR--SPA work as confirming the problem space while highlighting exactly the algebraic gap that geometric algebra fills.

\paragraph{Conceptual Spaces and Conceptual Semantics.} G\"{a}rdenfors's conceptual spaces framework \citep{gardenfors2000} shares the geometric intuition: meanings are regions in metric spaces, and composition involves geometric operations on those regions. \citet{gardenfors2012} have partially formalized composition for actions and events using conceptual spaces. However, the framework lacks a full algebraic type system, provides no obvious mechanism for type coercion or invertible transformation, and  treats composition primarily as prototype averaging or convex combination rather than structured algebraic combination. More broadly, \citet{jackendoff2025} have recently argued that word meanings draw on intricate combinations of material from a number of independent domains, each with its own computational affordances,'' distinguishing algebraic semantic structure (SemS) from analog spatial structure (SpS) and acknowledging that treating spatial structure in classical algebraic terms appears unpromising. FGA addresses both concerns: it provides the compositional algebra that Conceptual Spaces lacks, and it provides the single framework -- natively geometric \emph{and} algebraic -- that Jackendoff and Erk's multi-domain analysis calls for. The subspace decomposition $V = V_E \oplus V_P \oplus V_R \oplus V_S \oplus V_C$ is, I believe, a direct formal realization of their independent domains with their own computational affordances,'' and the geometric product is the unifying operation that links them compositionally without requiring stipulative cross-domain rules.

\paragraph{Vector space semantics for spatial language.}
\citet{zwarts1997vectors} and \citet{zwartswinter2000} place vectors directly into
the model-theoretic ontology: locative prepositions denote \emph{sets of vectors}
representing figure--ground positions, and constraints on those sets -- monotonicity,
conservativity -- parallel the semantic universals of Generalized Quantifier Theory.
\citet{zwarts2005prepositional} extends this to directional prepositions via an
algebra of paths, connecting spatial semantics to event structure.  This work
demonstrates that taking geometric structure seriously in the semantic metalanguage
reveals inference patterns invisible in a purely set-theoretic framework.  The
limitation is that the underlying algebra remains $\mathbb{R}^n$: no geometric
product, no grade structure, no rotors.  Position and distance are representable,
but not oriented transformations between reference frames, and the composition
operations do not generalize beyond the spatial domain.  FGA retains the core
insight -- vectors belong in the semantic ontology -- while upgrading the algebra
from linear to Clifford.

\paragraph{Modal Logics of Vector Spaces.} An independent line of support comes from modal logic, where there is a sustained tradition of treating vector spaces and their substructures as logical objects in their own right. For example, \citet{vanbenthem2024} show that vector spaces are not merely convenient containers for embeddings but support a rich internal logic: the set-level operations on groups validate the basic substructural logic of product and implication (i.e., the commutative Lambek Calculus), providing an independent argument that these spaces natively support compositional operations with the character of function application and relational binding. They also formalize the notions of linear dependence and independence as logical operators, capturing in modal terms exactly the subspace structure that FGA's type system exploits. A key finding is that the compositional principles they identify hold in vector spaces but fail in plain additive groups, because full field operations (including division) are required -- a formal argument for why geometric algebra, with its richer algebraic structure, provides compositional power that purely additive vector models cannot match. Relatedly, \ \citet{baltag2021} formalize functional dependence -- determination of one variable by others -- as a logical operator at a more abstract level, and FGA can be seen as bridging the algebraic dependence that van Benthem and Bezhanishvili study and the functional dependence that Baltag and van Benthem formalize. This program arrives at vector-space semantics from a direction entirely different from distributional NLP or cognitive geometry, and its convergence on the same structures reinforces the claim that the algebraic organization of vector spaces -- not merely their metric properties -- is what matters for compositional meaning.

\paragraph{Distributional Construction Grammars.} \ \citet{blache2024} have recently proposed integrating distributional semantics into Construction Grammar, motivated by the observation that comprehension involves both compositional mechanisms (incremental word-by-word assembly) and non-compositional ones (direct pattern recognition of form-meaning pairings). Their Distributional Construction Grammars embed distributional vectors within Sign-Based Construction Grammar feature structures, with an activation function selecting between compositional and direct-access routes. The diagnosis is consonant with what I have argued here: purely distributional methods lack compositional structure, but purely symbolic methods miss the continuous, usage-based information that distributional representations capture. Their solution is a hybrid architecture that juxtaposes the two. FGA offers a different path: a single algebraic framework in which compositional operations (contraction for application, wedge products for binding) and holistic pattern structures (blade-based representations at specific grades) coexist within the same Clifford algebra. Rather than switching between a symbolic and a distributional route, FGA provides one algebra in which both modes of meaning access are native operations.

% ============================================================
\section{Toward a Neural Implementation}
\label{sec:implementation}

A central question for any proposed semantic formalism is whether it can be learned, rather than merely stipulated. This question is particularly pressing for FGA, since the framework is motivated not only as a formal improvement over conventional linear algebraic semantics, but also as a representation suitable for neural learning. The issue, then, is not whether FGA can be embedded in a differentiable architecture in principle, but how its trainable parameters should be organized, what inductive biases they should encode, and how learning in such a system differs from the ordinary real-valued parameter learning familiar from large language models and other neural NLP systems.

\subsection{From flat vectors to graded lexical objects}

As developed in Sections~2--4, the FGA lexical parameter is not a flat vector $\mathbf{w} \in \mathbb{R}^d$ but a graded multivector in a Clifford algebra:
\begin{equation}
M_w = \sum_{k=0}^{K} \langle M_w \rangle_k, \qquad K \ll n \text{ in practice}.
\end{equation}
More explicitly, for a grade-truncated representation retaining grades $0$ through $2$, each lexical item $w$ is assigned parameters
\begin{equation}
M_w = s_w + \mathbf{v}_w + B_w,
\end{equation}
where $s_w \in \mathbb{R}$, $\mathbf{v}_w \in \mathbb{R}^n$, and $B_w = \sum_{i < j} b_{w,ij}\, e_i e_j$. The total number of trainable parameters per lexical item is $1 + n + \binom{n}{2}$ -- larger than a conventional embedding for the same base dimension, but the tradeoff is that the FGA object directly contains scalar, unary, and binary relational structure, whereas the LA embedding must rely on external learned maps to reconstruct those distinctions.

All coefficients $s_w, v_{w,i}, b_{w,ij}, \ldots$ are real-valued trainable parameters, but they are not interpreted as coordinates in a single homogeneous space. They are coordinates in a graded algebra in which the possible interactions between coordinates are constrained by the multiplication laws of the algebra itself. This changes what is learned where. In LA, one learns vectors and then separately learns how to compose them. In FGA, one learns multivector coefficients inside a representation whose algebra already supplies part of the compositional behavior. The representation carries not only semantic content, but also structured compositional potential.

\subsection{Why not simply learn larger LA vectors?}

An immediate objection is that the multivector parameterization appears to be nothing more than a larger real vector in disguise. Since a multivector in $\Cl(V,Q)$ can be coordinatized as a real vector of dimension $2^n$, why not say that FGA is simply a very large embedding space?

This objection misses the crucial difference between \emph{dimensionality} and \emph{organization}. A flat vector in $\mathbb{R}^{2^n}$ does not by itself distinguish which coordinates are scalar, which are bivector, which encode role structure, and which correspond to event-like interactions. By contrast, a Clifford algebra decomposes those coordinates into grades with fixed multiplication laws. The bivector basis element $e_i e_j$ is not an arbitrary index; it is the antisymmetric product of $e_i$ and $e_j$, and its behavior under multiplication is algebraically determined.

This distinction now has empirical support. The Geometric Algebra Transformer \citep[GATr;][]{brehmer2023} suggests that algebraic organization can matter independently of sheer dimensionality: by building transformer computation directly in a projective geometric algebra, it outperforms several non-geometric and equivariant baselines on tasks such as n-body modeling and robotic motion planning, while also showing that full transformer architectures can be implemented over multivector representations at practical scale.
 More recently, the Versor architecture \citep{hirst2026} extends this line of work by formulating sequence processing in conformal geometric algebra. Its design uses rotor-based state updates together with a geometric-product-based attention mechanism, and the reported results indicate gains over several transformer, graph, and GA baselines while using substantially fewer parameters than standard transformer models. These results are particularly relevant for FGA because they suggest that rotor-based state evolution---the same broad family of mechanism invoked here for coercion and semantic modulation---can serve as an efficient primitive for sequential computation.

CliffordNet \citep{ji2026}  provides complementary evidence from the vision domain. By organizing interaction around the Clifford geometric product rather than the familiar feed-forward-heavy design pattern, it suggests that richer algebraic interaction can reduce the need for brute-force parameter mixing. Ji's ablation result, in which the wedge-product component approaches the performance of the full geometric product, is especially noteworthy here, since it indicates that the antisymmetric bivector structure---precisely the part encoding relational orientation---carries a large share of the discriminative signal. 
All three architectures operate outside NLP, and I do not want to overstate the transfer; but the underlying principle -- that Clifford-algebraic structure provides an inductive bias superior to flat linear algebra -- strikes me as domain-independent.

There is also a cautionary data point from within NLP. The one existing test of GA specifically for word embeddings -- \citet{mani2016}'s Word2Mvec -- found mixed results: the geometric product caused coefficient explosion in a CBOW-style model, and skip-gram improvements over Word2Vec were modest. The lesson, I think, is that naive substitution of the geometric product into existing architectures is insufficient; the algebraic inductive bias must be deliberately aligned with the compositional structure of the task, as GATr's careful design of equivariant primitives and Versor's principled use of rotor accumulation demonstrate. The implementation strategy developed below is designed with this lesson in mind.

\subsection{Type-sensitive lexical priors}

Although one \emph{can} allow every lexical item to have fully dense coefficients across all retained grades, this ignores one of the main advantages of FGA: grade corresponds to semantic type. A better design imposes type-sensitive priors. Let $\tau(w)$ be the lexical type of item $w$. Then the trainable multivector may be written as
\begin{equation}
M_w = \mathcal{M}_{\tau(w)} \odot \Theta_w,
\end{equation}
where $\Theta_w$ is an unconstrained raw parameter tensor and $\mathcal{M}_{\tau(w)}$ is a type-conditioned mask or soft prior. For example, entities concentrate mass in grade~1, binary relations in grade~2, eventive predicates in grades~2--3. This provides regularization (reducing the hypothesis space to grade profiles that match the intended semantics), supports interpretability (if a token's learned mass concentrates in grade~2, this supports the claim that it functions as a relational operator), and comports directly with the formal commitments of the theory.

\subsection{Subspaces: doubly structured inductive bias}

Grade alone does not fully determine semantic type. As developed in Section~5, both entities and unary predicates occupy grade~1 but inhabit different subspaces; predicate bivectors and binding bivectors are both grade~2 but belong to different regions. The implementation must therefore preserve the decomposition
\begin{equation}
V = V_E \oplus V_P \oplus V_R \oplus V_C \oplus \cdots,
\end{equation}
so that lexical items are located by grade \emph{and} subspace support. The algebraic inductive bias is thus \emph{doubly structured}: grade encodes arity and type complexity, subspace encodes ontological category.

\subsection{Differentiable FGA operations}

Once lexical items are parameterized as multivectors, learning proceeds through differentiable algebraic operations. This is perhaps the most important implementation point: the move from LA to FGA does not block gradient learning. The geometric product, wedge product, left and right contraction, scalar projection $\langle M_u M_v \rangle_0$, and rotor action $R x \Rev{R}$ are all multilinear or polynomial in the underlying coefficients, hence differentiable almost everywhere and optimizable by standard first-order methods. The implementation challenge is not differentiability but representation design and inductive bias.

To be concrete about the backpropagation mechanics: the geometric product of two multivectors, when expressed in coordinates, is a bilinear map on their coefficient vectors. If $M_1$ has coefficient vector $\mathbf{c}_1 \in \mathbb{R}^{2^n}$ and $M_2$ has coefficient vector $\mathbf{c}_2 \in \mathbb{R}^{2^n}$, then the coefficient vector of $M_1 M_2$ is $\mathbf{c}_{\text{out}} = C(\mathbf{c}_1, \mathbf{c}_2)$ where $C$ is a bilinear form determined by the algebra's multiplication table. This is equivalent to a matrix multiplication $\mathbf{c}_{\text{out}} = G(\mathbf{c}_1)\, \mathbf{c}_2$, where $G(\mathbf{c}_1)$ is a matrix whose entries are linear functions of $\mathbf{c}_1$. Standard automatic differentiation (PyTorch, JAX) handles bilinear operations natively, and the chain rule applies without modification. The GATr \citep{brehmer2023}, Versor \citep{hirst2026}, and CliffordNet \citep{ji2026} implementations all use standard optimizers (Adam, AdamW) and achieve stable convergence on tasks ranging from n-body modeling to sequence classification. The optimization landscape has not been theoretically analyzed for Clifford-algebraic models specifically, but no evidence of pathological local minima has been reported in the existing implementations, and the algebraic constraints (grade structure, antisymmetry) may actually improve the loss landscape by reducing the effective dimensionality of the parameter space.

Three kinds of learnable parameters arise naturally:
\begin{itemize}
\item \textbf{Lexical multivector coefficients} -- the direct analogues of word embeddings: $(s_w, \mathbf{v}_w, B_w, T_w, \ldots)$ for each token $w$.
\item \textbf{Structural subspace parameters} -- the decomposition of the base space into $V_E, V_P, V_R$, either fixed or learned through orthogonality constraints.
\item \textbf{Compositional operators} -- grade weights for scoring, gates controlling grade participation, rotor parameters for contextual transformation, and projection maps for task heads.
\end{itemize}

In ordinary neural models, one learns vectors and then learns almost all compositional structure externally. In FGA, one learns multivector coefficients under fixed algebraic rules, and one may optionally learn additional control structure around those rules. The compositional core is partly algebraic rather than wholly architectural.

\subsection{Learning scenarios}

The simplest setting in which to test FGA is a lexical learning architecture analogous to a skip-gram or contrastive embedding model. Given token--context pairs $(w, c)$, one assigns multivectors $M_w$ and $M_c$ and defines a grade-sensitive score:
\begin{equation}
s_{\text{FGA}}(w,c) = \lambda_0\, \langle M_w M_c \rangle_0 + \lambda_2\, \|\langle M_w M_c \rangle_2\| + \lambda_3\, \|\langle M_w M_c \rangle_3\|,
\end{equation}
where the $\lambda_k$ are hyperparameters or learnable coefficients. The hypothesis is that if natural language contains relational and typed regularities that ordinary vector similarity collapses, then grade-sensitive objectives should learn lexical parameters that more faithfully encode those regularities.

A concern that should be addressed directly is whether FGA can be trained without extensive supervision. The rich semantic tasks described below (role labeling, entailment) require labeled data that is expensive to produce and often noisy. Several self-supervised objectives are available that exploit the algebraic structure of FGA without requiring semantic annotation:

\begin{itemize}
\item \textbf{Masked grade prediction.} Given a multivector representation of a word in context, mask one grade component (e.g., the bivector part) and train the model to reconstruct it from the remaining grades and surrounding context. This is the FGA analogue of masked language modeling, but it operates on grade components rather than tokens, and it exploits the algebraic constraints between grades as a training signal.
\item \textbf{Contrastive geometric learning.} For each observed compositional context (verb--argument pair, modifier--noun pair), construct a positive pair (the algebraically composed multivector) and negative pairs (compositions with randomly substituted arguments). Train the model so that the grade-0 score of the correct composition exceeds that of the corrupted compositions. This adapts the negative sampling strategy of skip-gram and contrastive learning to the grade-decomposed setting, exploiting the algebraic structure of correct versus incorrect compositions without requiring semantic labels.
\item \textbf{Grade-consistency as self-supervision.} The grade structure itself provides a training signal: if a token functions as an entity in a particular context, its learned representation should concentrate mass at grade 1; if it functions as a relation, at grade 2. Grade-consistency constraints -- penalizing representations whose grade profile does not match their syntactic function -- can be derived from unlabeled parsed text, providing a form of self-supervision grounded in the algebra's type system.
\end{itemize}

A richer setting uses FGA's compositional mechanics directly, with direct supervision from semantic tasks such as role labeling, event extraction, or entailment. The truth-conditional sentence score
\begin{equation}
s_t(\text{John loves Mary}) = j \lc (\ell \rc m) \in \Cl^0
\end{equation}
and the event-structural representation $\mathcal{E} = (r_{\textsc{ag}} \wedge j) + (r_{\textsc{th}} \wedge m)$ can both serve as training targets, yielding a composite loss:
\begin{equation}
\mathcal{L} = \mathcal{L}_{\text{truth}} + \beta\, \mathcal{L}_{\text{structure}} + \gamma\, \mathcal{L}_{\text{grade-reg}}.
\end{equation}
This differs from LA-based composition in a fundamental way: the operations of argument saturation, role binding, and type coercion are built into the compositional map rather than induced from data. Learning focuses on the lexical and operator parameters, not on rediscovering the basic mechanics of composition.

\subsection{Regularization}

A neural implementation faithful to FGA should regularize toward the structural commitments of the theory. Four regularizers suggest themselves naturally:
\begin{itemize}
\item \textbf{Grade sparsity}: encourage lexical items to concentrate mass in the grades appropriate to their semantic type.
\item \textbf{Subspace separation}: encourage learned entity, predicate, and role subspaces to remain distinct, via orthogonality penalties on their projection operators.
\item \textbf{Rotor normalization}: enforce $R\Rev{R} \approx 1$ for learned contextual rotors, through a penalty $\|R\Rev{R} - 1\|^2$.
\item \textbf{Type-consistency}: encourage derivations to end in the expected grade, penalizing mass in unexpected grades of compositional outputs.
\end{itemize}

\subsection{A realistic implementation path}

A full FGA-transformer is a natural long-term goal, but it is not the right starting point. The most realistic development path, motivated by both scientific caution and the lessons of prior work, is:
\begin{enumerate}
\item \textbf{Grade-truncated lexical model.} Learn token embeddings as $(s, \mathbf{v}, B)$ objects and evaluate whether the non-scalar grades acquire stable, interpretable semantic structure.
\item \textbf{Typed compositional learner.} Introduce contraction- and wedge-based sentence composition with supervision from semantic tasks.
\item \textbf{Contextual rotor layer.} Model contextual reinterpretation by explicit grade-preserving rotor transformations rather than opaque contextual embedding updates. A particularly informative experiment here would directly compare rotor-based contextual modulation with the standard projection-based mechanism ($Q$, $K$, $V$ matrices) on sense disambiguation tasks where the word's context selects among qualia-like aspects of its meaning -- the ``Romeo and Juliet'' versus ``Alfa Romeo'' cases that current systems handle through implicit subspace selection \citep{widdows2025}. The comparison would test whether the norm-preserving, invertible, grade-respecting properties of rotors offer advantages for compositional generalization that projections do not.
\item \textbf{FGA attention / sequence model.} Replace dot-product attention with grade-sensitive geometric interaction, as CliffordNet's sparse rolling mechanism demonstrates is computationally feasible at linear complexity, and as GATr \citep{brehmer2023} demonstrates for full multivector-valued hidden states.
\item \textbf{Signature adaptation.} Explore whether parts of the quadratic form can be learned once the lower-level components are stable. \citet{dehaan2024} provide the first systematic empirical comparison of Euclidean, projective, and conformal algebras within a single architecture, finding nontrivial tradeoffs that suggest signature selection is a genuine design parameter rather than a fixed choice.
\end{enumerate}

The central claim of FGA is not merely that one can replace one embedding formalism with another, but that the richer algebra captures semantic structure that flat vector spaces discard. The first empirical question is whether learned multivector lexical parameters actually realize this promise. Only after that is established does it make sense to scale to large contextual architectures.

\subsection{Summary: what FGA learning is really doing}

The basic answer to the implementation question is straightforward. Yes, FGA multivector parameters can be learned by ordinary gradient-based optimization, because the relevant algebraic operations are differentiable. The nontrivial issue is not learnability but representation design.

\begin{center}
\renewcommand{\arraystretch}{1.2}
\begin{tabular}{@{}lll@{}}
\toprule
 & \textbf{LA embedding} & \textbf{FGA embedding} \\
\midrule
Lexical parameter & vector in $\mathbb{R}^d$ & multivector in $\Cl(V,Q)$ \\
Native types & none & grades + subspaces \\
Similarity & dot product & scalar part of geometric product \\
Relational structure & external / learned & internal / grade-2 and above \\
Argument saturation & learned maps & contraction \\
Role binding & tensor / attention / ad hoc & wedge product \\
Contextual modulation & implicit contextualization & rotor action \\
Type checking & external & grade + subspace \\
Learning target & content in flat space & content in structured algebra \\
\bottomrule
\end{tabular}
\end{center}

That, I think, is the most explicit operational interpretation of the FGA program in neural terms: not abandoning learning, but replacing flat learnable vectors with learnable algebraic objects whose internal organization already reflects the kind of compositional semantics we want the model to support.

\paragraph{What exactly is learned, and what is the objective?}  To make the learning story fully operational, it is worth being explicit about three questions that the preceding discussion leaves implicit.

\emph{What parameters are learned?} Three kinds: (a) the multivector coefficients themselves -- the scalar, vector, bivector (and optionally higher-grade) components of each lexical item's representation; (b) the rotor parameters -- the bivector generators and angles that define contextual and qualia-based transformations; and (c) optional structural parameters such as grade weights $\lambda_k$, subspace projection matrices, and gate parameters controlling grade participation. In a grade-truncated model retaining grades $0$--$2$ with base dimension $n$, each lexical item has $1 + n + \binom{n}{2}$ trainable coefficients, and each rotor has $\binom{n}{2}$ bivector coefficients.

\emph{Can FGA learn a semantic parser?} In principle, yes: a semantic parser in FGA terms is a model that maps surface forms (words, phrases, sentences) to multivector representations such that the algebraic composition of the parts -- contraction for application, wedge products for role binding, rotors for coercion -- yields the correct representation of the whole. The learning objective for such a parser would include a \emph{structural loss} that penalizes deviations from well-formed compositional structures: if the parser assigns entity vectors $j, m$ and a predicate bivector $\ell$ to a transitive sentence, the composed result $j \lc (\ell \rc m)$ should be a scalar (grade 0) whose value matches the target truth-conditional judgment, and the event structure $(r_{\textsc{ag}} \wedge j) + (r_{\textsc{th}} \wedge m)$ should support correct role recovery. Misassignments -- such as placing an entity in grade 2, or a predicate in grade 0 -- would be penalized by the grade-consistency regularizer.

\emph{Does FGA offer a different training objective from standard embedding models?} This is perhaps the most important question, and the answer is yes. Standard embedding models -- skip-gram, masked language models, next-token prediction -- optimize \emph{predictive} objectives: learn representations that maximize the probability of observed contexts. FGA opens the possibility of an \emph{interpretive} objective: learn representations such that algebraic composition of subexpressions yields the correct representation of the whole. This amounts to a \emph{compositionality constraint}: the score of a well-formed derivation should be determined by the algebraic composition of its parts, not merely by the co-occurrence statistics of its surface forms. Operationally, one can define a compositional faithfulness loss:
\begin{equation}
\mathcal{L}_{\text{comp}} = \sum_{\text{derivations}} \left\| \text{Compose}_{\text{FGA}}(\text{parts}) - \text{target} \right\|^2 + \mu \sum_k w_k \cdot \text{grade-violation}_k,
\end{equation}
where the first term measures how well the algebraically composed result matches a target representation (from annotation, entailment supervision, or contrastive pairing), and the second term penalizes grade violations -- mass in grades where none should appear. The grade-sensitive skip-gram objective of Section~11.6 is a bridge to existing practice, using the familiar contrastive framework but with grade-decomposed scores. But the more distinctive contribution of FGA is this interpretive objective: rather than asking ``does this representation predict the next word?'' it asks ``does this representation compose correctly?'' -- whether the algebraic operations of contraction, wedge product, and rotor action, applied to the learned lexical parameters, produce the right compositional outcomes. This is the ``tightest structure'' intuition: the best representations are those for which the algebra does the most compositional work, leaving the least residual for the learning system to compensate for.

% ============================================================
\section{Conclusion}

I have argued in this paper that geometric algebra provides a mathematically richer representational framework than conventional linear algebra while maintaining full compatibility with vector learning methods. Functional Geometric Algebra pushes this further toward a unified theory of lexical semantics, compositionality, and inference.

The central claim is not merely increased dimensionality but increased \emph{structural organization}. GA expands an $n$-dimensional embedding space into a $2^n$ multivector algebra where base semantic concepts and their higher-order interactions are represented explicitly rather than approximated through parameter-heavy architectures. The geometric product subsumes and unifies the inner product (similarity), the outer product (relational binding), and the rotor sandwich (transformation) within a single algebraic operation. The grade structure provides an intrinsic type hierarchy. The three-layer FGA architecture -- contraction for truth-conditional semantics, wedge products for event structure, rotors for type coercion -- demonstrates that a single algebra can support the full range of semantic phenomena that currently require disparate mechanisms.

The evidence for this position is now converging from multiple directions simultaneously. On the engineering side, CliffordNet \citep{ji2026} demonstrates that the full geometric product -- both scalar coherence and bivector structure -- is so representationally dense that standard feed-forward networks become redundant, achieving state-of-the-art parameter efficiency in vision. If geometry is all you need for visual representation, I would argue that \emph{functional} geometry -- FGA -- is what you need for semantic representation, where the linguistic motivations for graded types, compositional binding, and type coercion provide even richer algebraic structure to exploit. On the architecture side, Clifford neural layers \citep{brandstetter2023}, Clifford equivariant networks \citep{ruhe2023}, and the Geometric Algebra Transformer \citep{brehmer2023} demonstrate that full multivector representations can serve as trainable hidden states within scalable deep learning architectures. On the positional encoding side, RoPE and CARE \citep{care2025} show that Clifford rotors already power modern transformers. On the embedding side, knowledge graph models \citep{xu2020,demir2023} demonstrate that the geometric product captures relational structure more naturally than flat vectors. And on the formal semantics side, DisCoCat's tensor contraction \citep{coecke2010} and DisCoCirc's dynamic state updates \citep{coecke2021} point toward the same algebraic operations that FGA unifies natively. From the direction of modal logic and algebraic logic, \citet{vanbenthem2024} independently demonstrate that vector spaces support rich modal-logical structure -- including an embedding of the Lambek Calculus and modal operators for linear dependence and independence -- arriving at the conclusion that the algebraic organization of vector spaces, not merely their metric properties, is what sustains compositional reasoning.

What remains is to bring these convergent threads together: to build and evaluate FGA-based neural architectures that exploit the full algebraic structure for natural language understanding. The mathematical foundations are in place -- a full formal development covering quantification, modality, discourse, and pragmatics within FGA is presented in \citet{pustejovsky2026fga}. The computational infrastructure is maturing -- CliffordNet's sparse rolling mechanism demonstrates that the geometric product can be computed at linear complexity, and GATr demonstrates that multivector-valued transformers scale to large problems. The linguistic motivations are clear. Geometric algebra is not just a theoretically elegant alternative to linear algebra for semantics -- it is, I believe, the right algebra for the job.

\section*{Acknowledgements}

I would like to thank the members of my lab for many fruitful debates on the compositional inadequacies of contemporary vector representations. In particular, I am especially grateful to Timothy  Obiso for discussions of hyper-dimensional vector encoding and holographic representations. I also  thank Kenneth Lai, Kyeongmin Rim, Bingyang Ye, and Yifan Zhu for their insights and useful  conversations on distributed representations more generally. I would also like to thank Vasanth Sarathy for  helpful and constructive comments on an earlier draft of the paper. I am particularly grateful to Dominic Widdows for detailed and generous comments on an earlier draft, which prompted significant clarifications of the grade-type correspondence, the relationship between FGA and existing transformer architectures, and the question of what explicit compositional structure adds beyond what large language models achieve implicitly. I also thank Janet Pierrehumbert for a thorough and incisive reading that sharpened the paper's claims about trainability, parameter complexity, and the relationship between FGA and contextual language models.
All errors are, of course, my own. 

\bibliography{GA_Position}

\end{document}